\documentclass[pdflatex,sn-vancouver-num]{sn-jnl}

\usepackage{xcolor}
\usepackage{soul}
\usepackage{graphicx}%
\usepackage{multirow}%
\usepackage{amsmath,amssymb,amsfonts}%
\usepackage{amsthm}%
\usepackage{mathrsfs}%
\usepackage[title]{appendix}%
\usepackage{xcolor}%
\usepackage{textcomp}%
\usepackage{manyfoot}%
\usepackage{booktabs}%
\usepackage{algorithm}%
\usepackage{algorithmicx}%
\usepackage{algpseudocode}%
\usepackage{listings}%
\usepackage{tikz}
\usepackage{bbding}
\usepackage{subcaption} 
\usetikzlibrary{arrows.meta}

\usetikzlibrary{positioning,fit,calc}
\usetikzlibrary{matrix}

\tikzset{
  block/.style={rectangle, draw, thick, minimum width=2.5cm, minimum height=0.75cm, align=center},
  circ/.style={circle, draw, thick, minimum size=5mm, inner sep=0pt},
  arrow/.style={->, thick},
}

\theoremstyle{thmstyleone}%
\theoremstyle{thmstyletwo}%

\theoremstyle{thmstylethree}%

\begin{document}

\title[Article Title]{Using Deep Learning Models Pretrained by Self-Supervised Learning for Protein Localization}


\author*[1]{\fnm{Ben} \sur{Isselmann}}\email{ben.isselmann@h-da.de}

\author[1]{\fnm{Dilara} \sur{G\"oksu}} \email{dilara.goeksu@stud.h-da.de}

\author[2,4]{\fnm{Heinz} \sur{Neumann}} \email{heinz.neumann@h-da.de}

\author[3]{\fnm{Andreas} \sur{Weinmann}} \email{andreas.weinmann@thws.de}

\affil*[1]{
  \orgdiv{Department of Mathematics and Natural Sciences},
  \orgname{Hochschule Darmstadt},%
  \orgaddress{
    \street{Schoefferstraße 3},
    \city{Darmstadt},
    \postcode{64295},
    \state{Hessen},
    \country{Germany}%
  }%
}
\affil[2]{%
  \orgdiv{Department of Chemical Engineering and Biotechnology},
  \orgname{Hochschule Darmstadt},%
  \orgaddress{
    \street{Stephanstrasse 7},
    \city{Darmstadt},
    \postcode{64295},
    \state{Hessen},
    \country{Germany}%
  }%
}
\affil[3]{%
  \orgdiv{Lab for Algorithms for Computer Vision, Imaging and Data Analysis},
  \orgname{Technische Hochschule Würzburg-Schweinfurt},%
  \orgaddress{
    \street{Münzstraße 12},
    \city{Würzburg},
    \postcode{97070},
    \state{Bayern},
    \country{Germany}%
  }%
}
\affil[4]{%
  \orgdiv{European University of Technology},
  \orgname{European Union}%
}



\abstract{\textbf{Background:} Task-specific microscopy datasets are often small, which makes it difficult to train deep learning models that learn robust and meaningful features. While recent advances in self-supervised learning (SSL) have shown promise, particularly through pretraining on large, domain-specific datasets, the generalizability of these models across datasets with differing staining protocols, imaging modalities, and channel configurations is not well explored. Yet this cross-domain transferability is crucial, especially for small-scale studies that cannot afford to generate large annotated datasets. Concretely, we conduct a systematic comparison of DINOv1-pretrained backbones across different pretraining sources and channel handling strategies, assessed by their ability to predict protein subcellular localisation patterns on the OpenCell dataset.

\textbf{Results:} DINOv1-based ViT backbones pretrained on either HPA FOV or ImageNet-1k transfer well to OpenCell, even without any fine-tuning, reaching a mean macro $F_1$ score of 0.689 $\pm$ 0.021 (17 classes, multi-class labels) in this zero-shot setting. Channel-wise embedding consistently achieved higher performance than channel replication. Fine-tuning on OpenCell further improved performance to a mean macro $F_1$ score of 0.704 $\pm$ 0.027 (17 classes, multi-class labels), while the gap between natural-image and microscopy-specific pretraining remained small throughout. At the single-cell level, embeddings from the HPA single-cell-pretrained backbone achieved the highest $k$-nearest neighbor performance across neighborhood sizes (macro $F_1$ score of 0.515 $\pm$ 0.040, 15 classes, multi-class labels), though a substantial portion of this reflects technical rather than biological similarity.


\textbf{Conclusion:} Self-supervised DINO backbones transfer effectively to fluorescence-microscopy protein localization, with natural-image pretraining nearly matching microscopy-specific pretraining and channel-wise embedding surpassing replication, offering practical guidance for reusing pretrained backbones on small microscopy datasets.
}



\maketitle

\section{Background}\label{sec1}


The analysis of human cells provides a crucial view into the molecular mechanisms underlying diseases and enables the identification of potential drug target references. Advances in high-throughput technologies have made it possible to investigate cellular function at an unprecedented scale \cite{jia_high-throughput_2022, navin_future_2011, buenrostro_single-cell_2015, bandura_mass_2009, wang_single-cell_2020}. In particular, transcriptomic profiling offers critical insights into gene function and regulation by analyzing RNA expression patterns, complementing genomics, which focuses on identifying mutations and structural variations \cite{heumos_best_2023}. However, these molecular approaches lack spatial resolution and morphological context. This limitation has led to the rise of image-based profiling, which enables the quantification of complex phenotypes and the assessment of treatment effects at the cellular level \cite{mattiazzi_usaj_high-content_2016, grys_machine_2016, scheeder_machine_2018}. Advances
in high-throughput microscopy and multiplex staining protocols now allow systematic study of subcellular
structures and protein localization across large-scale perturbation screens \cite{reicher_pooled_2024, vincent_phenotypic_2022, boyd_harnessing_2020}. 


Motivated to standardize image-based profiling, researches developed and still further develop the Cell Painting protocol \cite{gustafsdottir_multiplex_2013, bray_cell_2016}, a high-throughput staining method designed to extract rich, quantitative cell profiles. The stained and fixed cells are imaged across five microscopy channels. The recently optimized protocol \cite{cimini_optimizing_2023} highlights Cell Painting as the state-of-the-art assay for morphological profiling and provides a robust framework for the systematic analysis of cellular phenotypes. 

Inspired by the initial Cell Painting assay, both academia and the pharmaceutical industry generated several publicly available datasets. Most notable, the JUMP-CP consortium recently released the largest publicly accessible Cell Painting dataset to date, comprising images from over 116,000 chemical perturbations \cite{chandrasekaran_jump_2023}. While Cell Painting focuses on broad morphological profiling through organelle-level staining, other efforts have extended this idea toward protein-specific localization. Among them, the Human Protein Atlas (HPA) \cite{thul_subcellular_2017} adopts a more targeted approach using antibody-based immunofluorescence. Each HPA image includes three fixed reference channels: DAPI for the nucleus, $\beta$-tubulin as a cytoskeletal marker for microtubules, and calreticulin to label the endoplasmic reticulum (ER). These markers provide structural context for a fourth channel, which visualizes the protein of interest (POI) through antibody-based staining. In addition to the Human Protein Atlas, OpenCell \cite{cho_opencell_2022} is a large-scale imaging project with the goal of providing a high-confidence, artifact-free map of protein localization in human cells. Unlike antibody-based approaches, OpenCell uses CRISPR-mediated genome editing to endogenously tag proteins with fluorescent labels, enabling direct visualization of native protein distribution within cells. Alongside the protein of interest (POI), the nucleus is visualized using Hoechst 33342 staining in a second imaging channel.

A central challenge in image-based profiling is the generation of meaningful and robust feature representations. While traditional bio-imaging tools provide handcrafted features \cite{carpenter_cellprofiler_2006}, their interpretability comes at the cost of flexibility and scalability. Here, Deep learning (DL) offers a powerful alternative by learning task relevant representations directly from raw image data, reducing the need for manual feature engineering \cite{lecun_deep_2015, premkumar_single-cell_2024}. The use of deep learning models is often limited by their reliance on large annotated datasets, which are costly to generate and inherently restricted in their biological and experimental coverage. Recent studies have shown that self-supervised learning (SSL) can alleviate this bottleneck by learning transferable representations directly from unlabeled data \cite{kim_self-supervision_2025, doron_unbiased_2023}. Among them, Doron et al. \cite{doron_unbiased_2023} evaluated a self-supervised method known as DINO (self-distillation with no labels) \cite{caron_emerging_2021}, employing a Vision Transformer (ViT)~\cite{dosovitskiy_image_2021} as an encoder to extract biologically meaningful features across three publicly available imaging datasets with diverse biological focuses. In fluorescence microscopy, SSL has also been reported to reduce sensitivity to batch effects \cite{yao2024weaklysupervisedsetconsistencylearning}, promoting more generic and transferable feature representations. Beyond DINO, other SSL approaches such as masked autoencoders (MAE) have been proposed; for example, the study in \cite{kim_self-supervision_2025} compared several SSL models and demonstrated their effectiveness for morphological profiling and generalizability without fine-tuning, with only a small performance gap relative to supervised approaches. Recent efforts have focused on training more generic and transferable models that can accommodate heterogeneous channel configurations. Channel-adaptive approaches learn joint representations of multi-channel microscopy images by introducing channel embeddings and forming sequences of channel-aware patch tokens in ViT-based architectures \cite{bao_channel_2024, pham_cha-maevit_2025}. In contrast, channel-agnostic approaches expose the model to each channel separately during SSL training, aiming to learn representations that transfer across channel types without explicit channel conditioning \cite{lorenci_scaling_2025, agrawal_chammi-75_2025}. 


Despite these efforts and the demonstrated benefits of SSL, the practical reuse of pretrained models with a fixed input-channel configuration on small, task-specific datasets remains underexplored. In many scenarios, such datasets differ moderately from the pretraining distribution, yet share a subset of semantically related channels (e.g., nuclear markers such as DAPI or Hoechst, or channels encoding the protein of interest). For such settings, several concrete questions remain underexplored: (i) How do fixed-channel SSL-based ViT-backbones benefit from pretraining on large datasets (natural images vs. domain-specific fluorescence) with and without any fine-tuning? (ii) How can simple channel handling strategies resolve the channel mismatches? (iii) To what extent can an HPA single-cell–pretrained backbone provide meaningful single-cell embeddings when labeled data are scarce?

In this work, we investigate the extent to which fixed-channel DINOv1-based ViT backbones can be leveraged for protein localization tasks. Using the OpenCell dataset as a representative small, task-specific dataset and HPA FOV, HPA single-cell, and ImageNet-1k as pretraining sources, we systematically compare (a) backbones pretrained on large-scale datasets with and without fine-tuning on OpenCell, (b) channel replication versus channel-wise embedding for adapting mismatched channel layouts, , and (c) the ability of an HPA single-cell–pretrained backbone to provide meaningful embeddings on a small, expert-annotated subset of OpenCell single-cell samples. This study aims to provide simple, empirically grounded recommendations for how existing fixed-input SSL backbones can be leveraged most effectively for protein localization in fluorescence microscopy.
Building on our prior work \cite{isselmann2026generalizationselfsupervisedvisiontransformers}, which demonstrated strong cross-domain transfer without fine-tuning, we extend the analysis to fine-tuning, channel adaptation strategies, and single-cell evaluation.

\section{Methods}

\subsection{Datasets}
We used four DINO backbones to evaluate feature extraction on microscopic data. We trained One model on the OpenCell dataset, while the other three models were based on publicly available pretrained weights, one trained on ImageNet-1k and the other two on HPA dataset subsets from the corresponding Kaggle competition. To assess their performance, we used the OpenCell dataset as input to generate embeddings with each backbone, which were then used to predict protein localization. An overview of the datasets is provided in Table~\ref{tab:comparison_datasets}.

To represent a large and diverse set of natural images, we used a DINO backbone pretrained on ImageNet-1k \cite{russakovsky2015imagenetlargescalevisual}. ImageNet-1k is a well-established benchmark in computer vision, comprising over 1.2 million training images, 50,000 validation images, and 100,000 test images spanning 1,000 object categories. Originally developed for the ImageNet Large Scale Visual Recognition Challenge (ILSVRC), it is widely used for pretraining deep learning models for image classification. 

In the OpenCell \cite{cho_opencell_2022} dataset, the authors investigated the localization and interactions of human proteins to map the architecture of the human proteome. They generated a library of HEK293T cell lines by using CRISPR to insert fluorescent tags (split-mNeonGreen2) into 1,310 individual proteins. A total of 6,301 two-channel images (tagged protein and nucleus) were acquired. Protein localization was manually annotated across 17 subcellular compartments using a three-tier grading system: Grade 3 indicates prominent localization, Grade 2 denotes less pronounced localization, and Grade 1 represents weak localization patterns.

The HPA Cell Atlas provides two fluorescence microscopy datasets for protein localization. The HPA FOV dataset \cite{ouyang_analysis_2019} contains approximately 120,000 images from over 30 cell lines, combining 42,774 images from the Kaggle competition and an additional $\sim$~78,000 images from the HPAv18 dataset \cite{thul_subcellular_2017}, with four channels per image (protein, microtubules, nucleus, endoplasmic reticulum) and 28 expert-annotated localization classes. The HPA single-cell dataset \cite{le_analysis_2022} comprises roughly 105,000 images, including 23,589 from the Kaggle single-cell competition and $\sim$~82,000 from the HPAv20 dataset \cite{thul_subcellular_2017}, with 18 protein localization classes and one negative class, annotated at the compartment level across multiple cell lines. Both datasets were acquired using high-resolution confocal microscopy and provide detailed labels suitable for pretraining or evaluating models for protein localization.

\subsection{Channel handling methods}

\begin{figure*}[h]
  \centering

  \begin{subfigure}[t]{0.48\linewidth}
    \centering
    \includegraphics[width=0.75\linewidth]{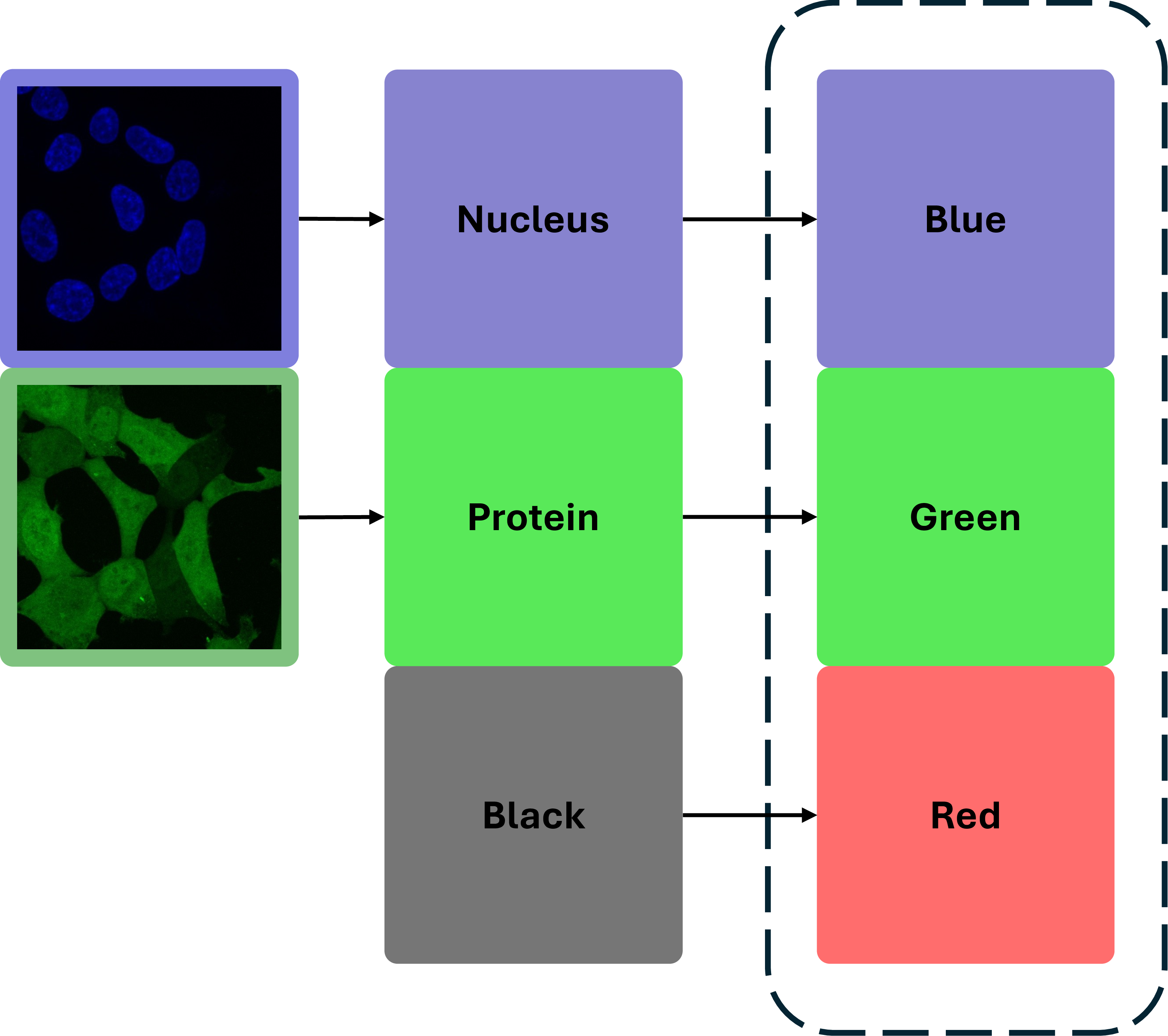}
    \caption{Channel-wise embedding of the ImageNet-1k pretrained backbone with three input channels}
    \label{fig:ch3map}
  \end{subfigure}\hfill
  \begin{subfigure}[t]{0.48\linewidth}
    \centering
    \includegraphics[width=0.75\linewidth]{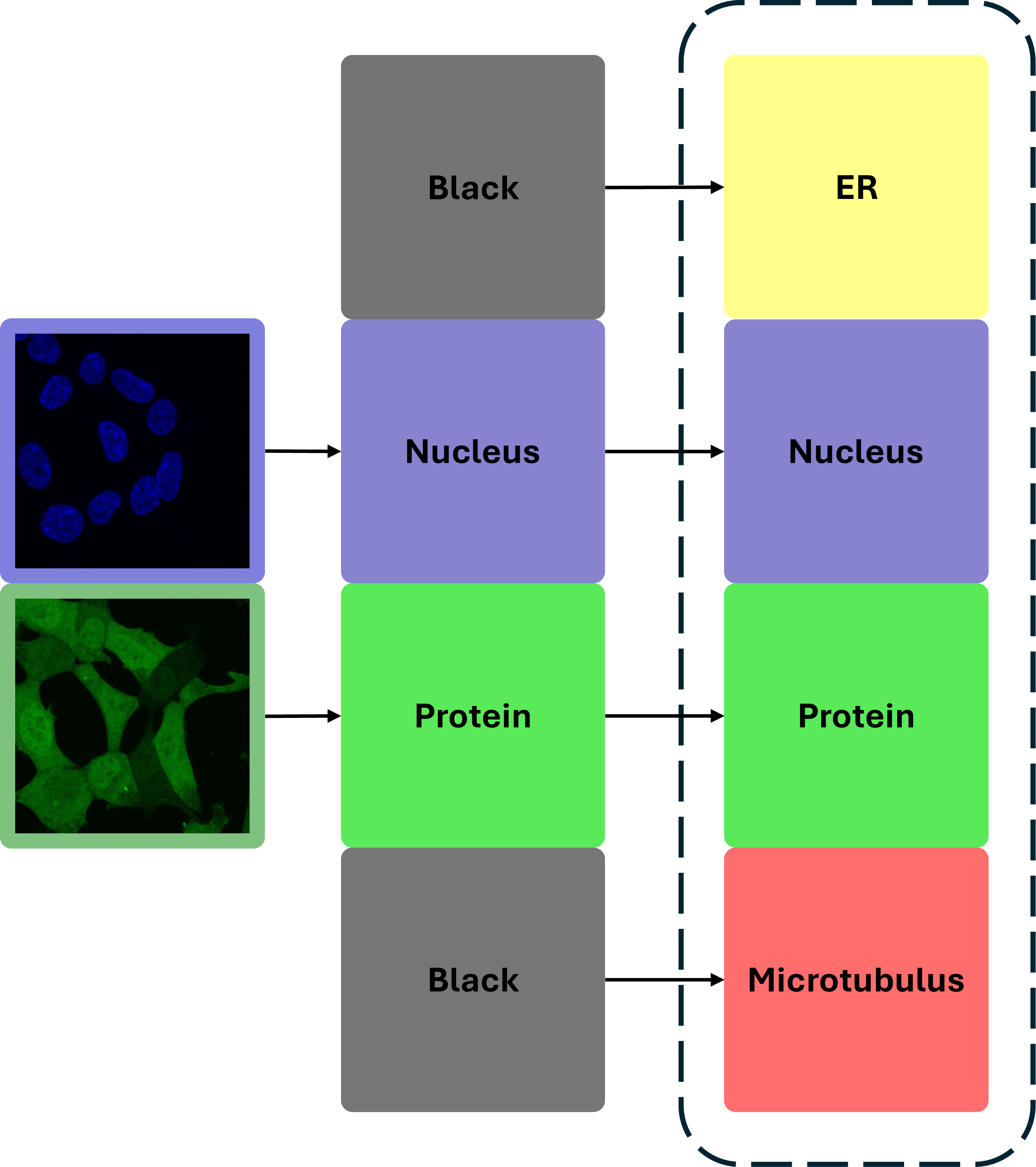}
    \caption{Natural channel-wise embedding of the HPA FOV pretrained backbone with four input channels}
    \label{fig:ch4map}
  \end{subfigure}

  \vspace{0.75\baselineskip}

  \begin{subfigure}[t]{0.48\linewidth}
    \centering
    \includegraphics[width=0.75\linewidth]{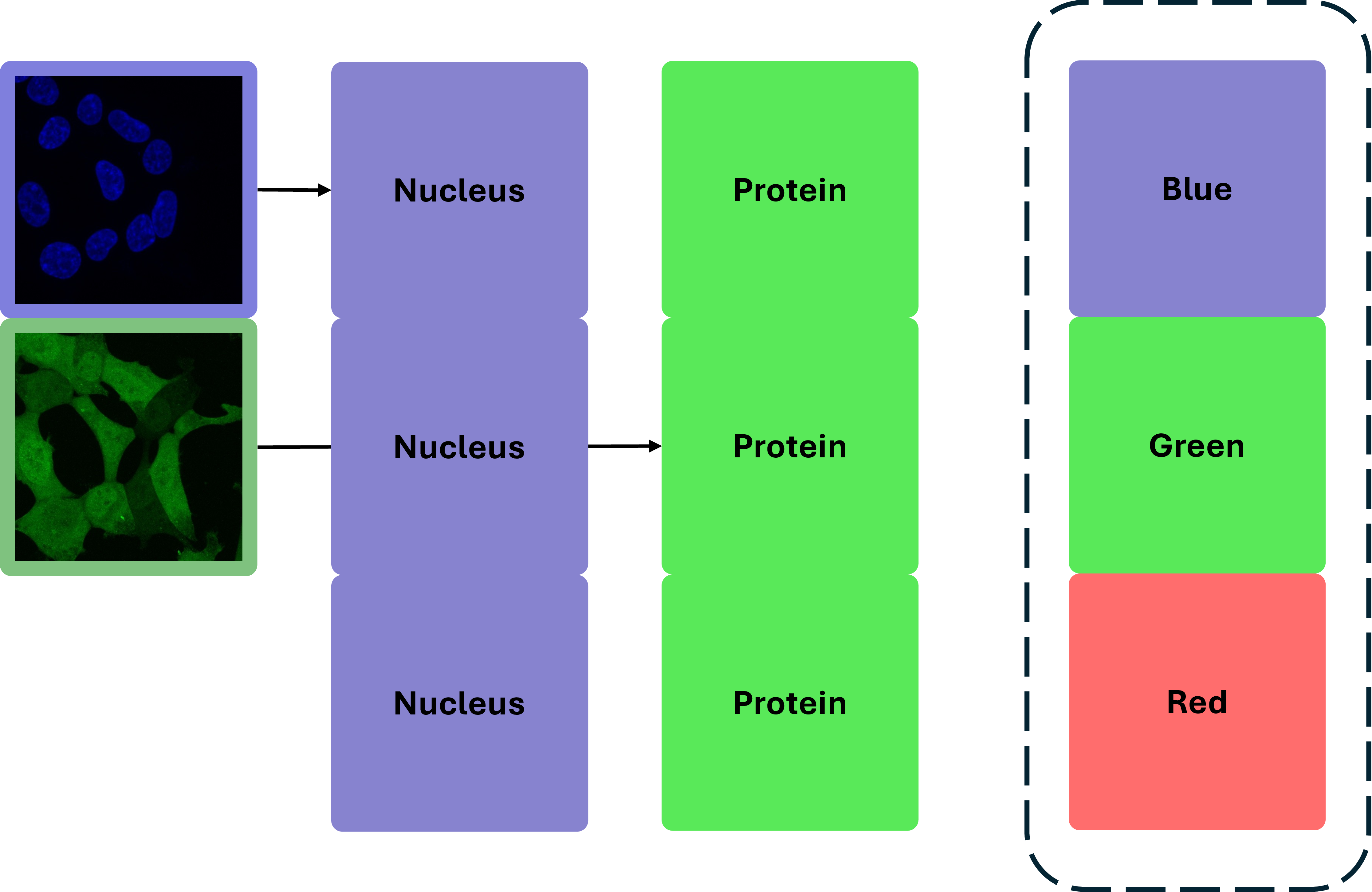}
    \caption{Channel replication of the ImageNet-1k pretrained backbone with three input channels}
    \label{fig:ch3rep}
  \end{subfigure}\hfill
  \begin{subfigure}[t]{0.48\linewidth}
    \centering
    \includegraphics[width=0.75\linewidth]{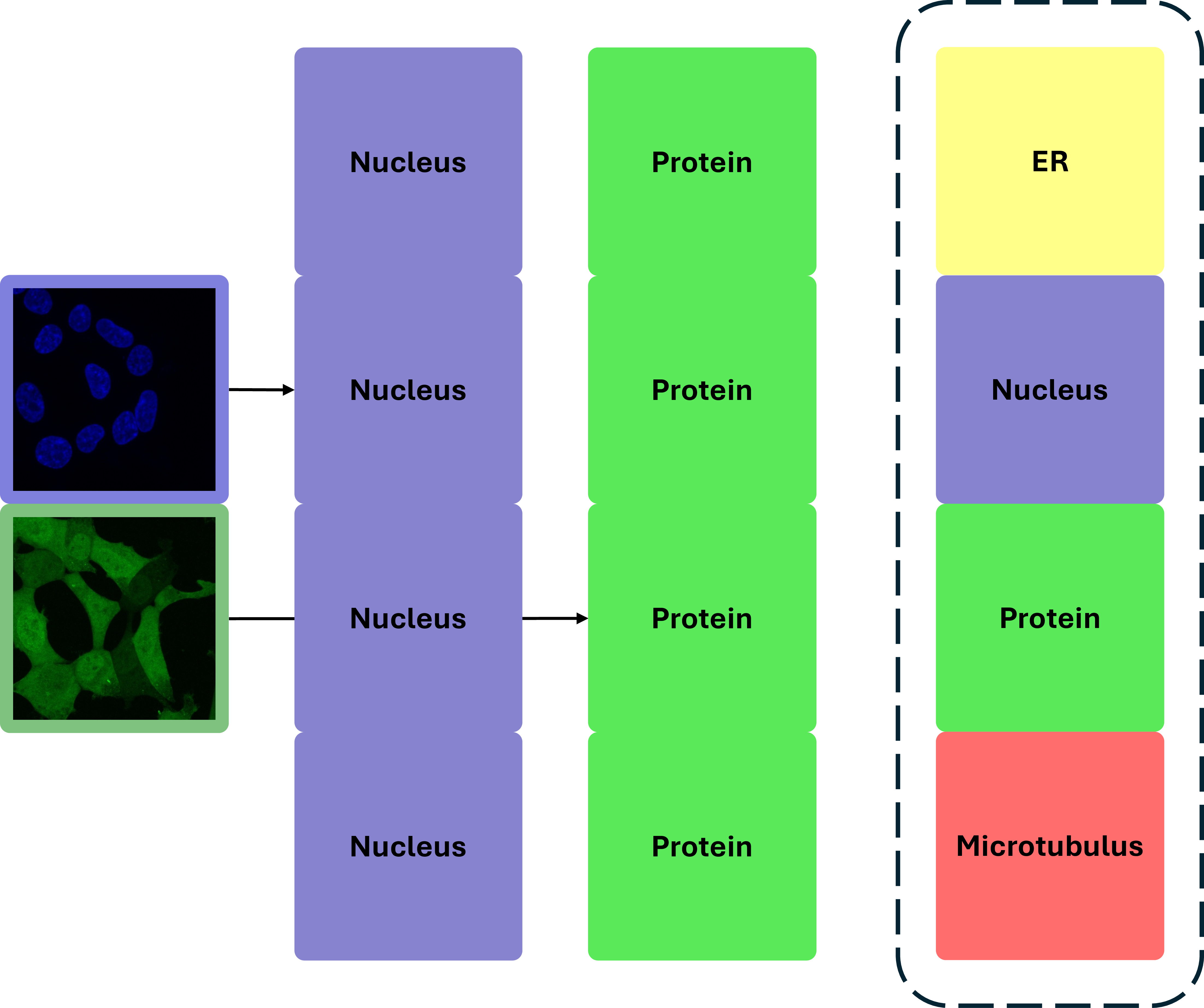}
    \caption{Channel replication of the HPA FOV pretrained backbone with four input channels}
    \label{fig:ch4rep}
  \end{subfigure}
  
  \caption{(Natural) channel-wise embedding (a,b) and channel replication (c,d) variants used to align microscopy inputs with pretrained DINO backbones.}
  \label{fig:channel-handling}
\end{figure*}

\label{subsec:embeddings}
To apply DINO models pretrained on ImageNet-1k and HPA to OpenCell data, it is essential to account for differences in channel composition across datasets. Since ViT backbones expect input formats consistent with their pretraining data, we explore two strategies to address this mismatch: Channel replication  \cite{Pawlowski085118, chen2024chammibenchmarkchanneladaptivemodels} and Channel-wise embedding.
\begin{enumerate}
    \item \textbf{Channel replication} We use the pretrained DINO model to extract features for each channel independently, as illustrated in Figure \ref{fig:channel-handling} (c,d) and \ref{fig:pipeline} b). Afterwards, we concatenate the individual feature vectors. The advantage is that no additional training is required. However, the computational cost and the dimensionality of the final feature vectors scale linearly with the number of channels \cite{chen2024chammibenchmarkchanneladaptivemodels, Pawlowski085118}.
    \item \textbf{Channel-wise embedding} We embed corresponding channels between datasets while padding missing channels with zeros. This approach benefits from the potential reuse of channel-specific features and requires no additional training. However, the effectiveness depends on the compatibility of the channel semantics between datasets. When handling the channel mismatch between OpenCell data and DINO pretrained on HPA, we channel-wise embed the protein and nucleus channels directly  (natural), while representing microtubules and ER with blank channels, as shown in Figure \ref{fig:channel-handling} (b). Similarly, for OpenCell to ImageNet-1k-pretrained DINO, we map the protein channel to the red (R) channel and the nucleus channel to the green (G) channel, as shown in Figure \ref{fig:channel-handling} (a). For both the OpenCell and HPA datasets, channels were indexed sequentially, such that the first channel (R) corresponds to index 0, the second (G) to index 1, and subsequent channels follow the same indexing scheme. Only the HPA datasets include a fourth channel, yellow (Y), with index 3. This indexing is used solely for clarity and consistency in the presentation of results.
\end{enumerate}

\subsection{DINO}
DINO \cite{caron_emerging_2021} is a self-supervised framework used to train an unbiased feature extractor.
The key idea is to enforce consistency between the outputs of a student and a teacher network, given different augmented views of the same image. This is achieved without the use of labels through a self-distillation objective.
Let $\theta_s$ be the parameters of a student network $g_{\theta_s}$ and $\theta_t$ the parameters of a teacher network $g_{\theta_t}$. Furthermore, let $x$ be an input image. Then, the student’s output probability distributions $P_s$ over $K$ dimensions is calculated by normalizing the output of the network $g$ with a softmax function:

\begin{equation}
P_s(x)^{(i)} = \frac{\exp(g_{\theta_s}(x)^{(i)}/\tau_s)}{\sum_{k=1}^{K}\exp(g_{\theta_s}(x)^{(k)}/\tau_s)}, \qquad 1 \leq i \leq K.
\end{equation}
Here, $\tau_s$ is a temperature parameter that controls the smoothness of the output distribution. The teacher's output $P_t$ is calculated analogously replacing $\theta_s$ by $\theta_t$ and $\tau_s$ by $\tau_t$.

Both networks share the same Vision Transformer (ViT) \cite {dosovitskiy_image_2021} architecture, which serves as the backbone for the DINO framework. The goal is to match the output of the student network $g_{\theta_s}$ to that of the teacher network $g_{\theta_t}$ by minimizing the cross-entropy loss with respect to the parameters of the student network $\theta_s$, i.e., minimizing $  H(P_t(x), P_s(x)) := - P_t(x) \log P_s(x)$, where on the right hand side, the $\log$ is applied component-wise, followed by the computation of a scalar product.  Specifically, a set of different views $V$ for the input image  $x$ is generated to obtain invariance to different augmentations. $V$ contains two global views $x_1^g, x_2^g$ and several local, smaller views. While the student processes every view in $V$, the teacher only sees the global views. This asymmetric setting prevents collapse and improves stability during training. The student's parameters $\theta_s$ are learned by minimizing 
\begin{equation}
\min_{\theta_s} \sum_{x \in \{x_1^g, x_2^g\}} \; \sum_{\substack{x' \in V \\ x' \neq x}} H(P_t(x), P_s(x'))
\end{equation}
using stochastic gradient descent. Here, the teacher's parameters $\theta_t$ are frozen. To update the weights of the teacher, an exponential moving average (EMA) on the student’s weights with the update rule 
\begin{equation}
    \theta_t \leftarrow \lambda \theta_t + (1 - \lambda)\theta_s
\end{equation}
was used where $\lambda$ follows a cosine schedule during training, see \cite{caron_emerging_2021} for details.

During training, both networks include a projection head on top of the backbone. This head is typically a multi-layer perceptron (MLP), and it maps the backbone’s output to a space where the self-distillation loss is applied. These heads are discarded after training. At inference time, we used only the backbone of the teacher network as a feature extractor.

\subsection{Training procedure}
\label{subsubsec:train_proc}

\begin{figure*}[h]
  \centering
  \includegraphics[width=0.95\linewidth]{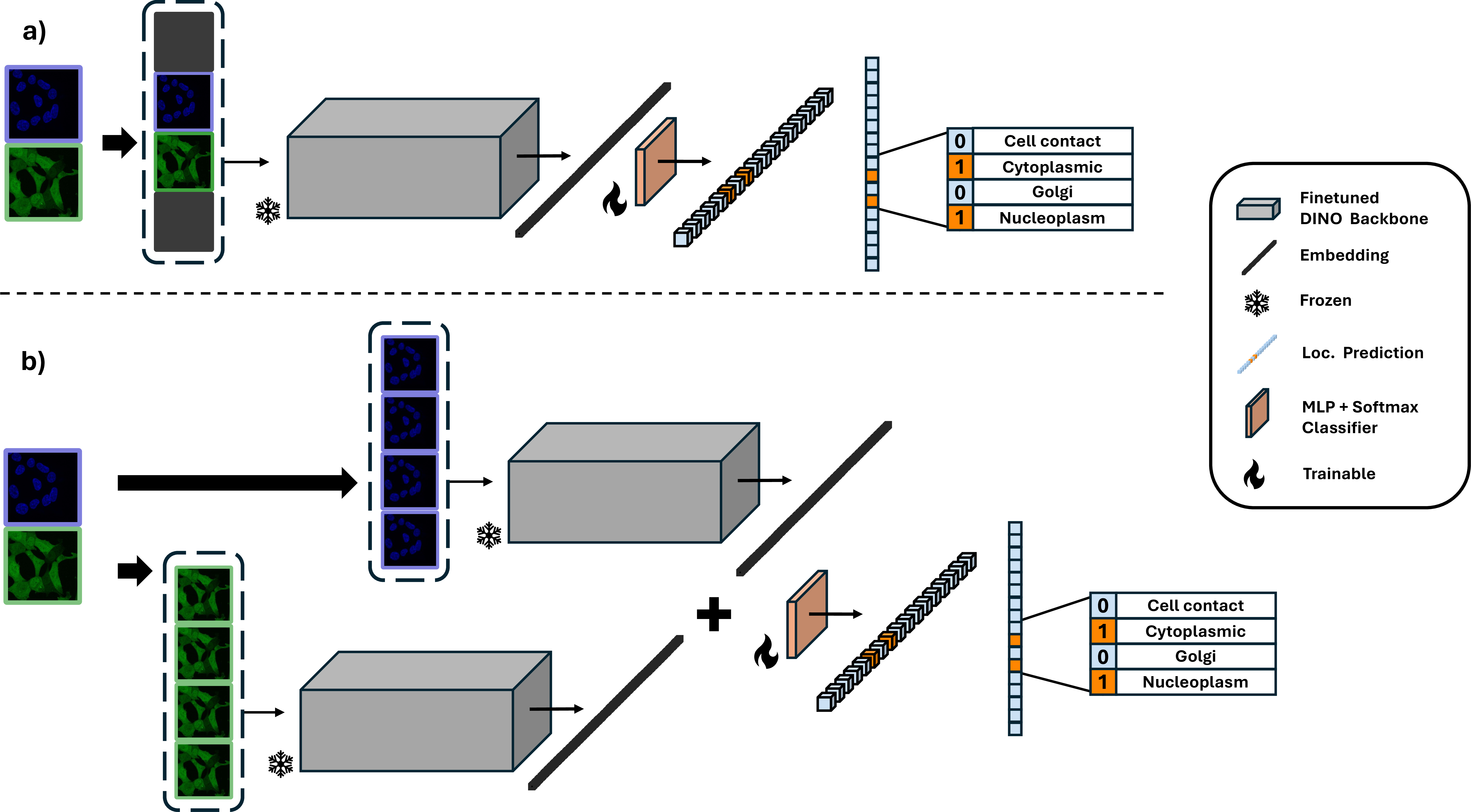}
  \vspace{4 mm}
  \caption{End-to-end pipeline for protein localization: OpenCell images are adapted to DINO-fine-tuned ViTs pretrained with different input-channel backbones (HPA FOV, ImageNet-1k, or OpenCell). Here, we illustrate a DINO-fine-tuned 4-channel model based on an HPA FOV backbone using (a) natural channel-wise embedding or (b) channel replication, after which a lightweight classification head is trained on OpenCell localization labels.}
  \label{fig:pipeline}
\end{figure*}

To predict protein localization labels for the OpenCell dataset, we employed the following two approaches:

\begin{enumerate}
    \item \textbf{FOV classification}: We adopted the two-stage approach proposed by Doron et al. In the first stage, we extracted feature embeddings from each OpenCell image using a (pre)trained DINO backbone. When additional fine-tuning on the OpenCell dataset was applied, this was explicitly indicated in the corresponding experiments. In the second stage, we trained a separate classifier head on each set of embeddings to predict the final protein localization labels. The training of the classifier to predict the protein localization is shown in Figure \ref{fig:pipeline} for OpenCell fine-tuning on HPA pretrained weights for the two channel handling strategies.
    \item \textbf{Single-cell classification}: Feature embeddings were extracted for individual segmented cells using a pretrained DINO backbone, with and without additional fine-tuning on the OpenCell dataset. To evaluate the quality of these embeddings, we applied a $k$-nearest neighbors classifier with varying neighborhood sizes ($k = 1, 3, 5, 10, 20$) to assess how well the learned representations capture subcellular localization patterns at the single-cell level.
\end{enumerate}
For details on the execution environment, see Appendix \ref{execution_environment}.
\subsection{Feature extraction with DINO}
\label{subsec:FeatWDino}
We explore four datasets for (pre)training DINO to generate embeddings of OpenCell images:
\begin{itemize}
    \item \textbf{Natural images (ImageNet-1k):} The DINO backbone pretrained on ImageNet-1k was originally trained in the initial DINO paper \cite{caron_emerging_2021}. The pretrained weights are publicly available in the corresponding GitHub repository (\url{https://github.com/facebookresearch/dino}).
    
    \item \textbf{Microscopic datasets (HPA FOV and HPA single-cell):} The authors in \cite{doron_unbiased_2023} trained various DINO backbones from scratch on a subset of the HPA dataset at both the FOV and single-cell level. In contrast to the original DINO setup for natural images with three RGB channels, they adapted the architecture to process four-channel fluorescence microscopy images. The pretrained weights are publicly available in the corresponding GitHub repository (\url{https://github.com/broadinstitute/Dino4Cells_analysis}).
    
    \item \textbf{Downstream task dataset (OpenCell):} We trained DINOv1-based ViT-backbone from scratch on the OpenCell dataset, as well as using the OpenCell dataset for fine-tuning the pretrained backbones. Images were processed through an augmentation pipeline including random resized crops (global: $224$px, scale $0.4$-$1.0$; local: $96$px, scale $0.05$-$0.4$), cell warping, and additional spatial transformations ($50\%$ flip probability). The ViT backbone weights were initialised directly from the respective pretrained DINO checkpoints~\cite{caron_emerging_2021}, following the architecture specification of Dosovitskiy et al.~\cite{dosovitskiy_image_2021}. The projection head architecture and weight initialisation follow the original DINO implementation~\cite{caron_emerging_2021}. For both training from scratch and fine-tuning, we used identical hyperparameters: 200 epochs, a cosine scheduler for learning rate ($lr = 3 \times 10^{-3}, lr_{min} = 1 \times 10^{-4}$), weight decay ($wd = 0.04, wd_{end} = 0.4$), and teacher momentum ($\lambda = 0.996, \lambda_{end} = 1$),  center momentum $m = 0.9$, 20 warm-up epochs, an effective batch size of 48 across 4 GPUs (12 per GPU). The model was optimized with an AdamW optimizer \cite{loshchilov2019decoupledweightdecayregularization}. For self-supervised pretraining, we trained on $90\%$ of the available data, holding out $10\%$ as a test set for evaluation.
\end{itemize}

\subsection{Evaluation}
\label{subsec:eval}
We employed different prediction strategies to evaluate the backbones and channel-handling methods. Backbones trained at the FOV level were assessed using a softmax classifier, whereas, due to the scarcity of expert-labeled single-cell data, single-cell embeddings were evaluated with a $k$-nearest neighbor classifier.

\subsubsection{Classifier}
Starting with the (pre-)trained DINO backbones, which are based on the four datasets described in the previous Section~\ref{subsec:FeatWDino} - and which were used exclusively as frozen encoders in this downstream task, i.e., their weights remain frozen throughout classifier training - we first derived latent representations of all input images. Each embedding feature was standardized to have zero mean and unit variance using statistics computed on the training set; the same normalization parameters were then applied to the validation and test sets. To address class imbalance, we resampled each training example with a probability inversely proportional to the frequency of its rarest label. Then, we constructed a simple classification head to evaluate their performance on the OpenCell dataset.

A multi-layer perceptron (MLP) classifier consists of three layers:
\begin{enumerate}
\item \textbf{Input linear layer:} 512 hidden units, ReLU activation, dropout regularization ($p = 0.5$);
\item \textbf{Hidden linear layer:} 256 hidden units, ReLU activation, dropout regularization ($p = 0.5$);
\item \textbf{Output linear layer:} 17 output units corresponding to the protein localization classes.
\end{enumerate}

The classifier weights and biases were initialized using PyTorch's default initialization scheme and applied to all three MLP layers. Parameters were sampled from a uniform distribution $U(-1/\sqrt{n},\, 1/\sqrt{n})$, where $n$ denotes the number of input features to the layer \cite{paszke2019pytorch}. During training, the the classifier head was optimized using the AdamW optimizer (weight decay~$=~0.04$, $\beta_1 = 0.9$, $\beta_2 = 0.999$) with an initial learning rate of $10^{-4}$. We trained for 300 epochs with a batch size of 512, employing a cosine annealing learning rate scheduler. The loss function used was binary cross-entropy with logits.

To ensure robust evaluation, we conducted five-fold cross-validation on the entire OpenCell dataset, grouping the images by gene to guarantee that no gene appears in more than one fold.

\subsubsection{Evaluation using $k$-nearest neighbor}
To directly assess the discriminative quality of the learned DINO embeddings without additional supervised training, we employed a $k$-nearest neighbor evaluation in a leave-one-out setting. Each feature vector from the OpenCell dataset was treated as a query and compared against all remaining embeddings using cosine similarity.

For each query, the top-$k$ most similar neighbors were identified, and their multi-label annotation vectors were retrieved. We adopt soft voting as in DINO \cite{caron_emerging_2021}. In this setting, each neighbor is assigned a weight given by:
\begin{equation}
w_i = \exp\left(\frac{s_i}{\tau}\right).
\end{equation}
Here, $s_i$ denotes the cosine similarity between the query and its $i$-th neighbor, and $\tau~=~0.07$ is a temperature parameter controlling the sharpness of the weighting distribution. The final class scores for each query were computed as the weighted sum of neighbor label vectors, followed by per-sample normalization to the range $[0,1]$. Since many proteins in the OpenCell dataset localize to multiple compartments, we convert the continuous class scores into binary predictions by applying a threshold of $0.5$.

\subsubsection{Evaluation metrics}

Because we solved a multi-label classification problem, where each data point can be assigned to zero, one, or several non-overlapping classes, we used the macro-averaged $F_1$ score \cite{jurafsky2025speech} as the evaluation metric. First, we state the notions of precision and recall for each class $i$ which were defined as 

\begin{equation}
\text{precision$_i$} = \frac{\text{tp$_i$}}{\text{tp$_i$} + \text{fp$_{i}$}}, \qquad \text{recall$_{i}$} = \frac{\text{tp$_i$}}{\text{tp$_i$} + \text{fn$_{i}$}}.
\end{equation}

Here, $\text{tp}_i$ represents the number of true positives for class $i$, while $\text{fn}_i$ and $\text{fp}_i$ denote the numbers of false negatives and false positives, respectively.
Next, the $F_1$ score for class $i$ is defined as the harmonic mean of precision and recall, i.e., 
\begin{equation}
F_{1i} = 2 \cdot \frac{\text{precision$_i$} \cdot \text{recall$_i$}}{\text{precision$_i$} + \text{recall$_i$}}.
\end{equation}
Finally, the macro-averaged $F_1$ score is computed as the arithmetic mean of the $F_1$ scores across all $n$ classes: 
\begin{equation}
\text{macro $F_{1}$} = \frac{1}{n}\sum_{i=1}^{n} F_{1i}.
\end{equation}

The annotated benchmark for the $k$-nearest neighbor evaluation contains only 61 genes and each gene corresponds to a single field of view (see Appendix~\ref{single-cell-masks-create}). We therefore quantify the variability of the aggregate macro-F1 by bootstrapping at the gene level rather than the cell level. We created 1000 bootstrap replicates, resampling the 61 genes with replacement, pooling all single cells belonging to the resampled genes, and recomputing the macro-F1 score (averaged over the 15 present localization classes) on each replicate. The $k$NN predictions themselves are computed once on the full annotated set and held fixed across replicates, so the resulting spread reflects sampling variability in the gene population — i.e. how much the score would be expected to vary over a different draw of genes. We report the mean and standard deviation of this bootstrap distribution as mean $\pm$ SD in Table~\ref{tab:knn-results}. We do not report per-class confidence intervals: several classes are represented by as few as two genes, so per-class interval estimates would be unreliable, and we treat the per-class scores (see Table~\ref{tab:per-class-f1}) as descriptive.






\section{Results}
\label{results}
Motivated by the question of whether large domain-specific datasets such as HPA and cross-domain datasets such as ImageNet-1k can be leveraged as pretraining sources for small, task-specific datasets like OpenCell, we systematically fine-tuned these pretrained models on the OpenCell dataset and quantify the resulting gains in protein localization performance at field-of-view (FOV) and single-cell level.

\subsection{FOV protein localization prediction}
\label{subsec:res_FOV}
We first performed experiments at the field-of-view (FOV) level to assess the transferability of cross-domain and domain-specific pretrained models under fine-tuning. In particular, we evaluated the effect of fine-tuning on different pretrained backbones, studied how the size of the task-specific fine-tuning set influences performance, and compared random channel-wise embeddings against embeddings that respect the distinct, naturally corresponding channels used in domain-specific pretraining datasets. To evaluate our results, we followed the training strategy described in Section \ref{subsubsec:train_proc}. For a clearer overview, Tables~\ref{tab:vit-opencell},~\ref{tab:vit-imagenet},  and~\ref{tab:vit-hpafov} report solely mean macro $F_1$ scores; the extended tables containing full metrics are provided in Appendix \ref{appendix:table1_2}.

\subsubsection{DINOv1 training on OpenCell from scratch}
To evaluate self-supervised learning with DINOv1 when feature representations are learned from scratch on OpenCell, we first compared two ViT backbones, ViT-small/8 and ViT-base/8, with results reported in Table~\ref{tab:vit-opencell}. Although ViT-base/8 has substantially more parameters, ViT-small/8 achieves the higher mean macro $F_1$ score (0.686 $\pm $0.026), suggesting that smaller backbones are beneficial when training DINOv1 on limited data.

\begin{table}[h]
\caption{Experimental results for ViT models trained 
from scratch on OpenCell. All models perform inference on the OpenCell dataset. 
Values are reported as mean $\pm$ one standard deviation.}\label{tab:vit-opencell}
\begin{tabular*}{0.7\textwidth}{@{\extracolsep{\fill}}lc@{}}
\toprule
Model & Mean Macro $F_1$ \\
\midrule
ViT-small/8 & 0.686 ($\pm$ 0.026) \\
ViT-base/8  & 0.631 ($\pm$ 0.021) \\
\botrule
\end{tabular*}
\end{table}

\subsubsection{Zero-shot and fine-tuning on ImageNet-1k backbones}
We further investigated the cross-domain generalizability of ImageNet-1k pretrained backbones in both zero-shot and fine-tuning setting, and additionally examined the channel-handling strategies introduced in Section~\ref{subsec:embeddings}. The results are reported in Table~\ref{tab:vit-imagenet}. All ImageNet-1k backbones transfer to OpenCell even without fine-tuning, reaching a mean macro $F_1$ scores $\geq$ 0.653 $\pm 0.018$ and thereby demonstrating cross-domain generalizability. Notably, the highest zero-shot performance is competitive with the domain-specific model trained from scratch on OpenCell (see Table \ref{tab:vit-opencell}). While the ViT-base/8 backbone performs best in the zero-shot setting, the ViT-small/8 backbone again shows an advantage when fine-tuned on the limited OpenCell data, achieving the overall highest performance among the ImageNet-1k backbones (mean macro $F_1$ score of 0.701 $\pm$ 0.023). For both backbones, channel-wise embedding tends to achieve slightly higher $F_1$ scores than channel replication; interestingly, channel replication decreases performance when fine-tuned on OpenCell, whereas channel-wise embedding benefits from fine-tuning.
\begin{table}[h]
\caption{\label{tab:vit-imagenet}Experimental results for ImageNet-1k 
pretrained backbones evaluated on OpenCell, without (No FT) and with (FT) 
fine-tuning. Values are reported as mean $\pm$ one standard deviation.}
\begin{tabular}{l l c c}
\toprule
Model & Channel Handling Method & No FT & FT \\
\midrule
ViT-small/8 & Channel-wise embedding & 0.665 ($\pm$ 0.018) & 0.701 ($\pm$ 0.023) \\
ViT-small/8 & Channel replication & 0.653 ($\pm$ 0.018) & 0.633 ($\pm$ 0.020) \\
\midrule
ViT-base/8  & Channel-wise embedding & 0.688 ($\pm$ 0.022) & 0.695 ($\pm$ 0.020) \\
ViT-base/8  & Channel replication & 0.674 ($\pm$ 0.023) & 0.639 ($\pm$ 0.030) \\
\botrule
\end{tabular}
\end{table}

\subsubsection{Zero-shot and fine-tuning on the HPA FOV backbone}
\label{res:generalizability}
Motivated by the question of whether domain-related ViT backbones can further improve embeddings for protein localization prediction, with and without fine-tuning, we investigated the HPA FOV backbone. Given the findings from the two previous experiments (Tables~\ref{tab:vit-opencell} and~\ref{tab:vit-imagenet}) regarding the potential benefit of the ViT-small/8 backbone, we note that this configuration was not pretrained by the authors of~\cite{doron_unbiased_2023}; we therefore report results only for ViT-base/8 in Table~\ref{tab:vit-hpafov}.
The HPA FOV backbone with natural channel-wise embedding achieves a mean macro $F_1$ score of 0.682 $\pm$ 0.034 in the zero-shot setting, improving to 0.705 $\pm$ 0.029) after fine-tuning on OpenCell --- the highest fine-tuned performance across all evaluated backbones. As observed for the ImageNet-1k backbones, channel-wise embedding consistently outperforms channel replication, and this advantage is preserved both with and without fine-tuning. Fine-tuning improves both channel-handling strategies for the HPA FOV backbone, in contrast to the ImageNet-1k setting where channel replication degraded under fine-tuning.
\begin{table}[h]
\caption{\label{tab:vit-hpafov}Experimental results for HPA FOV pretrained 
backbones evaluated on OpenCell, without (No FT) and with (FT) fine-tuning. 
Values are reported as mean $\pm$ one standard deviation.}
\begin{tabular}{l l c c}
\toprule
Model & Channel Handling Method & No FT & FT \\
\midrule
ViT-base/8 & Natural channel-wise embedding & 0.682 ($\pm$ 0.034) & 0.705 ($\pm$ 0.029) \\
ViT-base/8 & Channel replication            & 0.611 ($\pm$ 0.031) & 0.640 ($\pm$ 0.037) \\
\botrule
\end{tabular}
\end{table}
  


\subsubsection{Natural channel-wise embedding vs.\ random channel embedding}
\label{res:channelwise_emb}
We further investigated channel-specific transferability by comparing our best configuration --- natural channel-wise embedding on the HPA FOV-pretrained backbone (see Figure~\ref{fig:channel-handling}) --- against baselines using random channel embeddings in the zero-shot setting. Deviations from the natural channel alignment are denoted by the mapping order; for example, $[1,2] \rightarrow [2,1]$ indicates that the protein and nucleus channels are swapped relative to their naturally corresponding positions. This design tests whether the specific channel embedding, rather than an arbitrary input ordering, drives performance.
As shown in Table~\ref{tab:vit-hpa-results-random-no-additional-epochs}, the most influential factor is whether the protein channel is embedded naturally: configurations preserving the natural protein alignment achieve the highest macro $F_1$ scores ($\geq 0.663 \pm 0.030$). The best performance is obtained when both the protein and nucleus channels are embedded to their naturally corresponding positions (macro $F_1$ score of 0.681 $\pm$ 0.033). When only the nucleus is naturally aligned, performance drops to mean macro $F_1$ score 0.634 $\pm$ 0.028, and when both channels are mapped to unrelated inputs, the highest score reaches only 0.625 $\pm$ 0.028.

In Appendix~\ref{channel-embedding-appendix}, we provide an extended channel-wise embedding study detailing how natural channel alignment influences downstream performance both in the zero-shot setting and when the HPA FOV backbone is fine-tuned on OpenCell. While the effect is clear in the zero-shot baseline, the performance gap largely diminishes after fine-tuning on OpenCell.

 \begin{table}[h]
\caption{\label{tab:vit-hpa-results-random-no-additional-epochs}Experimental results for a ViT pretrained on HPA FOV evaluated on OpenCell. Values are reported as mean $\pm$ one standard deviation.}

\begin{tabular}{l l l c c}
\toprule
Pretrained Dataset & Model & Natural & Channel-wise embedding & Mean Macro $F1$ \\
\midrule
HPA FOV & ViT-base/8 & Yes &  [1, 2] $\rightarrow$ [1, 2] & \textbf{0.681 $\pm$ 0.033} \\
HPA FOV & ViT-base/8 & No &  [1, 2] $\rightarrow$ [1, 3] & 0.666 $\pm$ 0.028 \\
HPA FOV & ViT-base/8 & No &  [1, 2] $\rightarrow$ [1, 0] & 0.663 $\pm$ 0.030 \\
HPA FOV & ViT-base/8 & No &  [1, 2] $\rightarrow$ [0, 2] & 0.634 $\pm$ 0.028 \\
HPA FOV & ViT-base/8 & No &  [1, 2] $\rightarrow$ [3, 0] & 0.625 $\pm$ 0.028 \\
HPA FOV & ViT-base/8 & No &  [1, 2] $\rightarrow$ [0, 1] & 0.621 $\pm$ 0.025 \\
HPA FOV & ViT-base/8 & No &  [1, 2] $\rightarrow$ [2, 1] & 0.609 $\pm$ 0.021 \\
\botrule
\end{tabular}
\end{table}

\subsection{Single-cell protein localization prediction} \label{sc}

Here, we provide a more detailed view of how well different (pre)training setups, across pretraining datasets, backbone sizes, and patch resolutions, capture biologically meaningful single-cell structure  in a zero-shot setting. To assess the quality of the learned single-cell embeddings, a subset of the generated single-cell masks was labeled for correct protein localization by an expert, with the procedure described in detail in Appendix \ref{single-cell-masks-create}. The resulting benchmark comprises 992 single cells drawn from 61 genes (each from a distinct field of view) and covers 15 localization classes, with at least two genes per class (median four, range two to ten). As channel handling method, we only used channel-wise embedding as described in Section \ref{subsec:embeddings}. We examined the zero-shot embedding quality of each backbone via a $k$-nearest neighbor classifier across different neighborhood sizes $k$. To prevent influence of technical field-of-view- and gene-level information on the retrieval, we applied a leave-one-out strategy in which all cells sharing the query cell's gene (and therefore its field of view) were excluded from the neighbor pool during inference. Results are shown in Table~\ref{tab:knn-results}.

\begin{table}
\caption{Leave-one-out kNN macro-F1 by backbone. The query cell's entire gene
is excluded from the neighbour pool, so matches cannot exploit gene or FOV identity.
Macro-F1 is averaged over the 15 present localisation classes; the final column gives the
standard deviation over 1000 gene-level bootstrap replicates (near-constant across $k$).}
\label{tab:knn-results}
\centering
\begin{tabular}{l l ccccc c}
\toprule
Pretraining & Model & $k{=}1$ & $k{=}3$ & $k{=}5$ & $k{=}10$ & $k{=}20$ & $\pm$SD \\
\midrule
HPA FOV & ViT-B/8 & 0.369 & 0.405 & 0.407 & 0.411 & 0.405 & 0.03 \\
HPA FOV ft. OpenCell & ViT-B/8 & 0.419 & 0.435 & 0.437 & 0.440 & 0.437 & 0.04 \\
HPA single-cell & ViT-B/16 & \textbf{0.496} & \textbf{0.505} & \textbf{0.506} & \textbf{0.515} & \textbf{0.501} & 0.04 \\
ImageNet-1k & ViT-B/16 & 0.426 & 0.453 & 0.459 & 0.458 & 0.446 & 0.03 \\
ImageNet-1k & ViT-B/8 & 0.432 & 0.440 & 0.442 & 0.446 & 0.441 & 0.03 \\
\botrule
\end{tabular}
\end{table}

The best $k$-nearest neighbor performance was obtained using embeddings from the HPA single-cell pretrained ViT-base/16 backbone, which achieved the highest macro-averaged $F_1$ score consistently across all evaluated neighborhood sizes, with a peak of macro $F_1 = 0.515$ at $k = 10$. A per-class breakdown (Table~\ref{tab:per-class-f1}, Figure~\ref{fig:confusion-matrices}) shows this aggregate advantage is not uniform: the HPA single-cell backbone's gains are concentrated in nuclear and nucleolar classes — nucleolus GC/FC-DFC, chromatin, and nuclear membrane — where it both attains the highest per-class F1 and roughly halves the cross-confusion between the two nucleolar subclasses relative to the ImageNet-pretrained backbones. Since macro-F1 weights all classes equally, this reflects sharper discrimination of these specific patterns rather than an effect of class frequency. Small-support classes (centrosome, cytoskeleton, cell contact) remain weak across all backbones, consistent with the limited number of genes per class and contributing to the overlap in aggregate scores. Interestingly, the zero-shot HPA FOV backbone yields the weakest embeddings across all values of $k$, even marginally below both ImageNet-1k backbones, suggesting that the structural gap between FOV-level and single-cell images limits the transferability of FOV-level representations to the single-cell regime. Fine-tuning the HPA FOV backbone on the OpenCell FOV dataset closes this gap, bringing it on par with the ImageNet-1k backbones and indicating that domain-specific microscopy pretraining tends to remain beneficial even when the pretraining and inference image scales differ. In general, larger values of $k$ tend to yield higher classification performance.


\subsection{Qualitative analysis of protein localization predictions}
In addition to the quantitative evaluation, we conducted a qualitative analysis to better understand how the model represents protein localization and where it succeeds or fails. Specifically, we inspected the latent space of the best-performing model (HPA FOV–pretrained, OpenCell-fine-tuned backbone with channel-wise embedding) and visually examined input images with their predicted localization patterns. 

\begin{figure*}[!t]
    \centering
    \includegraphics[width=\linewidth]{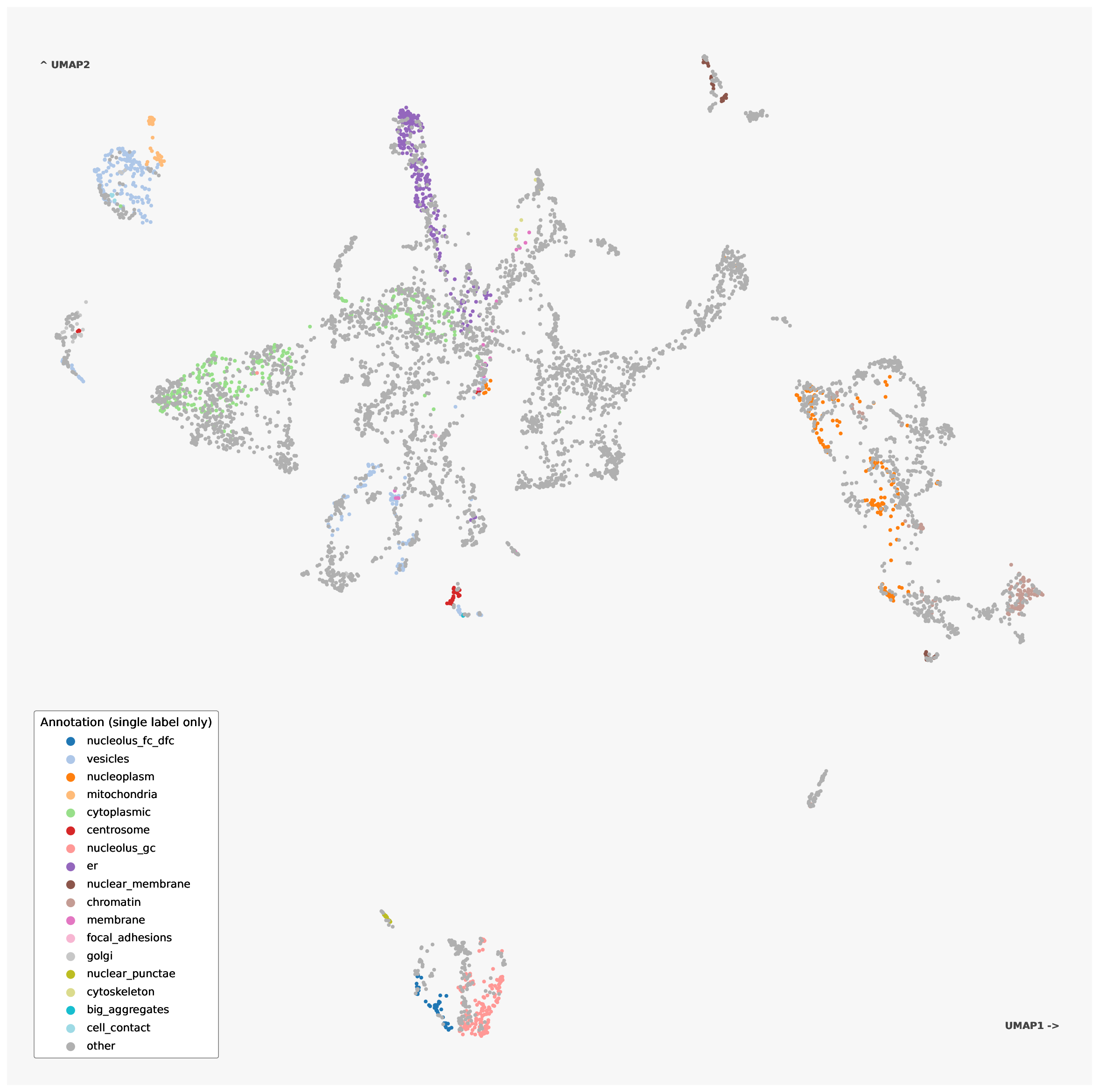}
    \caption{
        UMAP projection of OpenCell embeddings produced by the best-performing model (HPA FOV–pretrained, OpenCell-fine-tuned backbone with natural channel-wise embedding). 
        Each point corresponds to a single-cell embedding and is colored according to its protein localization label. We color proteins with one specific label only.
    }
    \label{fig:umap_embeddings}
\end{figure*}

Figure~\ref{fig:umap_embeddings} shows a UMAP projection of the OpenCell embeddings produced by this model. We colored only those embeddings whose proteins are annotated with a single subcellular localization, while embeddings with multiple localizations are shown in gray. The projection indicates that the model is able to capture protein localization patterns, particularly in the case of single localization, as evidenced by well-formed clusters and clear separation between distinct localization groups. 

\begin{figure*}[h]
    \centering
    \includegraphics[width=\linewidth]{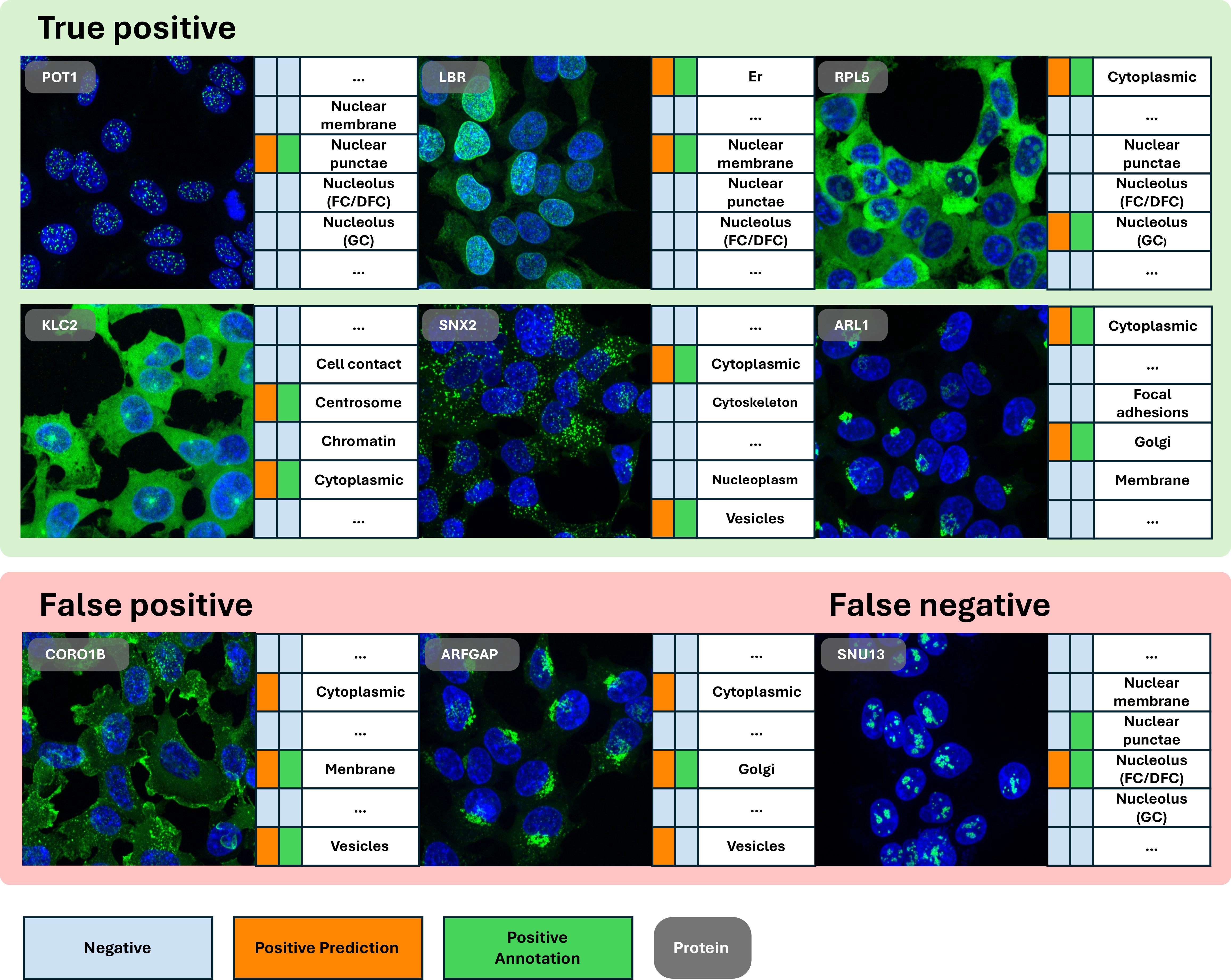}
    \caption{
    Representative qualitative examples from the best-performing model (HPA FOV–pretrained, OpenCell-fine-tuned backbone with channel-wise embedding). 
    For each example, we show the input image, indicate the tagged protein, and list the predicted protein localization together with the reference label(s). 
    The examples illustrate correctly classified prototypical patterns (true positives, TP) as well as challenging cases with ambiguous, mixed, or weak localization signals. 
    }
    \label{fig:qual_examples}
\end{figure*}

In addition, we examined representative image samples, as shown in Figure~\ref{fig:qual_examples}. For each example, we display the input image, the tagged protein together with the corresponding predicted protein localization pattern, and the reference label(s). This allows us to highlight both prototypical cases, where predictions align well with the annotations, and challenging cases, where discrepancies between labels, input images, and model outputs become apparent. Notably, several examples that are counted as false positives nevertheless show biologically plausible signals in the predicted compartment, such as clear cytoplasmic staining, dispersed vesicles that may have budded from the Golgi, or additional nuclear puncta beyond the nucleoli. In these cases, the model appears to pick up weak or subtle structures that were not captured in the original weak labels. More generally, we observed that apparent errors frequently arise when complementary localization signals overlap at the same or closely adjacent pixels (e.g. Golgi and vesicles, or nuclear puncta and nucleolus, FC/DFC) or when the fluorescence signal of a given compartment or organelle is weak relative to its surroundings, which also makes expert annotation inherently difficult.

\section{Discussion}\label{sec12}

We discuss the implications of the findings in Section~\ref{results} for using SSL-pretrained ViT backbones and outline key limitations.

Both ImageNet-1k and HPA FOV pretrained backbones transfer well to OpenCell, and neither pretraining domain clearly dominates. While the domain-related HPA FOV backbone achieves the nominally highest fine-tuned performance, its advantage over the natural-image ImageNet-1k backbone remains within the fold-level variability. This suggests that large-scale natural-image pretraining already provides a strong and transferable feature basis for protein localization, and that domain-specific pretraining offers, at most, a modest additional benefit.
Both backbones also benefit from fine-tuning on OpenCell, though the magnitude of this benefit varies. The ImageNet-1k backbone shows the largest improvement from the zero-shot to the fine-tuned setting, consistent with a larger domain gap that fine-tuning helps to close. While the smaller backbone tended to perform favorably when fine-tuning on limited data, we refrain from drawing a general conclusion about backbone size, noting only that smaller backbones remained competitive and are worth considering when data is limited.
Turning to channel handling, channel-wise embedding consistently achieves higher performance than channel replication across all direct comparisons — that is, within the same backbone and setting — and this gap widens further after fine-tuning for both the ImageNet-1k and HPA FOV backbones. Beyond the choice between embedding and replication, the specific channel assignment also matters: in the zero-shot setting, natural channel-wise embedding provides a marked benefit, particularly when the protein channel is mapped to its naturally corresponding pretrained channel. This indicates that the channel assignment, rather than an arbitrary input ordering, carries meaningful information for transfer. Notably, this advantage largely diminishes after fine-tuning, suggesting that fine-tuning can compensate for a suboptimal channel assignment.

Beyond FOV-level prediction, we examined the quality of single-cell embeddings using a small expert-annotated subset of the OpenCell dataset, evaluated with a $k$-nearest neighbor classifier. To ensure that retrieval reflects biological similarity and to rule out potential impacts of technical similarity, we adopted a leave-one-out strategy excluding all cells sharing the query cell's gene, and therefore its field of view, thereby removing shared field-of-view-level structure such as batch or plate effects as a potential confounder.
Under this evaluation, two observations emerge. First, zero-shot embedding quality is influenced not only by the pretraining data domain but also by the scale of the pretraining images relative to the inference images: backbones pretrained at the FOV level transfer tends less effectively to single-cell representations than those pretrained directly on single-cell crops. Second, despite the limited number of labeled single cells and the use of a non-parametric classifier, embeddings from SSL-pretrained backbones, particularly the HPA single-cell model, show a weak but consistent biological signal, concentrated in nuclear and nucleolar localizations rather than distributed uniformly across classes (Section~\ref{knn-appendix}), though the margins between backbones lie within the bootstrap standard deviation and absolute differences should therefore not be over-interpreted. This indicates that domain-specific SSL pretraining at single-cell granularity can yield useful feature extractors even when only very limited single-cell labels are available.
At the same time, the small size and weakly supervised nature of the single-cell benchmark impose clear limitations: absolute performance estimates should be interpreted with caution, and more extensive, high-quality single-cell annotations will be needed to validate and extend these observations.

In addition to the quantitative evaluation, we qualitatively inspected the SSL-based embeddings to assess their utility for downstream protein localization prediction. As shown in the UMAP projection in Figure 5, the learned representations exhibit well-separated, coherent clusters that broadly correspond to distinct subcellular localization classes — notably, nucleus-related and cytoplasm-related labels occupy clearly distinct regions of the latent space. Importantly, this structure emerges without any localization supervision during pretraining, suggesting that the self-supervised objective implicitly captures morphological features that are informative for protein localization.

Closer inspection of individual prediction examples (Figure 6) reveals that several false positive compartments exhibit plausible protein signals, suggesting that the model may 
detect weak or subtle localization patterns that were not captured by the original weak labels. This raises the possibility that the learned representations generalize beyond the granularity of expert annotation — a characteristic consistent with the label transcendence observed in the UMAP structure (Figure 5). At the same time, cases where the model struggles to distinguish signals from adjacent compartments reflect a fundamental biological ambiguity: overlapping fluorescence signals from multiple subcellular structures within the same channel represent an inherent limit for both computational models and human annotators alike.

Although our results demonstrate the benefits of SSL-based backbones for extracting biologically meaningful features in fluorescence microscopy, such as protein localization patterns, we also emphasize that these findings come with several important limitations that constrain the scope of this study. First, we relied on two microscopy datasets only. Our downstream dataset, OpenCell, comprises a single cell line imaged with one specific microscopy setup, and the HPA data are derived from a single overall assay. As a result, the diversity of imaging domains covered in this study is limited. Second, this study focuses exclusively on protein localization and does not consider more general phenotypic profiling tasks or perturbation responses. Including such endpoints in future work would allow for more general and broadly applicable recommendations on how to use SSL-pretrained backbones for small, task-specific microscopy datasets. In the scope of this study we only used DINOv1-based backbones, but it should be mentioned that there are other SSL approaches such younger versions of DINO \cite{oquab_dinov2_2024, simeoni_dinov3_2025}, MAE \cite{he_masked_2021}, and SimCLR \cite{chen_simple_2020} not assessed in this work. Lastly, the single-cell masks used in this work rely on Cellpose, which can introduce segmentation errors that directly affect the extracted features and, consequently, the downstream evaluation of single-cell embeddings.

\section{Conclusion}\label{sec13}

In this work, we investigate to what extent fixed-channel SSL-pretrained ViT backbones can be leveraged for protein localization tasks in fluorescence microscopy, using OpenCell as a representative small, task-specific downstream dataset to examine localization prediction, and using HPA FOV, HPA single-cell, and ImageNet-1k as pretraining sources. Our results show that the prediction performance of SSL-based ViT backbones benefits from large-scale pretraining even without any fine-tuning, and that the performance gap between natural-image and domain-specific microscopy pretraining is small in the zero-shot setting. When fine-tuning the backbones on OpenCell, both pretraining sources converge to a similar level of performance; while the domain-specific HPA FOV backbone retains a marginal advantage, this final gap remains within fold-level variability, indicating that the choice of pretraining domain has only a modest effect on downstream performance. Regarding channel handling, which is a key obstacle when reusing fixed-channel backbones, channel-wise embedding tends to perform more favorably than the replication strategy. Importantly, this advantage is most pronounced when downstream channels are aligned with their naturally corresponding channels in the pretrained source. 


Finally, results on a small, expert-annotated subset of OpenCell single-cell masks evaluated with a $k$-nearest neighbor classifier (Section~\ref{sc}) suggest that SSL pretraining at the single-cell level can capture a weak but genuine biological signal. Future SSL-based single-cell feature extractors must therefore focus on emphasizing biological information rather than capturing technical confounders.

This work encourages the use of fixed-channel SSL-based backbones for both FOV-level and single-cell datasets, provided that known relationships between channels in the downstream and pretraining datasets are explicitly exploited. Building on these insights, future work should explore more channel-agnostic ways of reusing fixed-channel backbones across heterogeneous microscopy assays and integrating more diverse datasets and more recent SSL strategies.

\backmatter

\section*{Declarations}

\bmhead{Supplementary information}
Not applicable.

\bmhead{Acknowledgements}
Not applicable.


\bmhead{Funding}
HN and AW acknowledge funding by the German Research Foundation (DFG) under grant INST 168/4-1.

\bmhead{Ethics declaration}
Not applicable.

\bmhead{Consent for publication}
Not applicable.

\bmhead{Consent to publish declaration}
Not applicable.

\bmhead{Consent to Participate declaration}
Not applicable.

\bmhead{Data availability}
All data used during this study are available via the OpenCell AWS Open Data Registry (\url{https://registry.opendata.aws/czb-opencell}). The repository was accessed on 1 May 2025.
All quantitative analyses (model training and inference) were performed on unaltered raw images; contrast adjustments were applied only for visualization in Figure \ref{fig:qual_examples}.

\bmhead{Competing interests}
The authors declare that they have no competing interests.

\bmhead{Code availability}
The code is publicly available at \url{https://code.fbi.h-da.de/microscopy/dino4opencell}.

\bmhead{Author contribution}
BI: Study design, DNN training, experiments, analysis, manuscript preparation. 
DG: DNN training, experiments, analysis, manuscript preparation. 
HN: Data labeling, manuscript revision, funding acquisition.
AW: Study design, manuscript revision, supervision, funding acquisition.


\begin{appendices}

\section{Further information on datasets}
Table \ref{tab:comparison_datasets} provides additional details on the datasets used in this work, comparing ImageNet-1k with OpenCell, HPA FOV, and HPA single-cell.
\begin{sidewaystable}
\caption{\label{tab:comparison_datasets}ImageNet-1k, OpenCell, HPA Image Classification and HPA single-cell Classification datasets.}
\begin{tabular}{p{0.195\linewidth}p{0.2\linewidth}p{0.275\linewidth}p{0.275\linewidth}}
\toprule
& ImageNet-1k \cite{russakovsky2015imagenetlargescalevisual} & OpenCell \cite{cho_opencell_2022} & HPA FOV \cite{ouyang_analysis_2019} / HPA single-cell \cite{le_analysis_2022}\\
\midrule
Images & $\sim$ 1.2 million & 6,301 & $\sim$ 120,000 / $\sim$ 105,000 \\
Number Labels & 1000 & 17 & 28 protein localization labels / 18 protein localization labels and one negative label for negative or unspecific signal  \\
Labels description & object categories & Protein localization labels available for 6120 images, (grades 1–3), Single-protein localization  per cell (OpenCell ontology) & expert-annotated localization classes, Compartment-level  annotations per image~/~cell \\
Cell lines & - & 1 (only HEK293T cell line) & over 30 different cell lines \\
Domain & Natural images & Fluorescence microscopy & Fluorescence microscopy  \\
Channels & 3 (RGB images) & 2 (protein, nucleus) & 4 (protein, microtubules, nucleus, endoplasmic reticulum) \\
Microscope & - & Spinning disk confocal  (Andor Dragonfly) &  Leica SP5 point-scanning confocal\\ 
Objective & - & 63× / 1.47 NA oil immersion & 63× / 1.40 NA oil immersion \\

Pixel size & -  & 0.102µm (102.4nm) & 0.08µm (80nm) \\

Lateral resolution  & - & $\sim$ 211nm & $\sim$ 222nm\\

Axial resolution  & - & $\sim$ 750–850nm & $\sim$ 785nm \\

Bit depth  & - & 16-bit & 16-bit\\

Environmental control  & - & Live imaging @ 37°C, 5~\%~CO$_2$ & Fixed cells (immuno-staining) \\

Cell state  & - & Live cells & Fixed, immuno-labeled cells  \\

Typical cells per image  & - & $\sim$ 10–30 cells per field & $\sim$ 10–30 cells per field  \\

Well format  & - & 96-well plates & Glass-bottom chamber slides or multi-well plates\\

Image size  & - & Typically 1024×1024 pixels, with cropped FOVs at 600×600 pixels & Typically 2048×2048 px \\

URL & \url{https://www.image-net.org/} & \url{https://opencell.czbiohub.org/} & \url{https://www.proteinatlas.org/} \\
Further information & ImageNet-1k consists of photographs gathered from multiple search engines, with labels assigned manually. & Out of the 6,301 FOV images, 181 did not have an assigned protein localization label. Only the labeled images were included in our study. & HPA FOV consists of the Kaggle dataset with approximately 42,774 images, along with around 78,000 images from the HPAv18 dataset. HPA single-cell includes 23,589 images from the Kaggle dataset and an additional 82,495 images from the HPAv20 dataset.  \\
\botrule
\end{tabular}
\end{sidewaystable}

\section{Extraction Single-Cell Masks} \label{single-cell-masks-create}
\begin{figure*}[h]
    \centering
    \includegraphics[width=\linewidth]{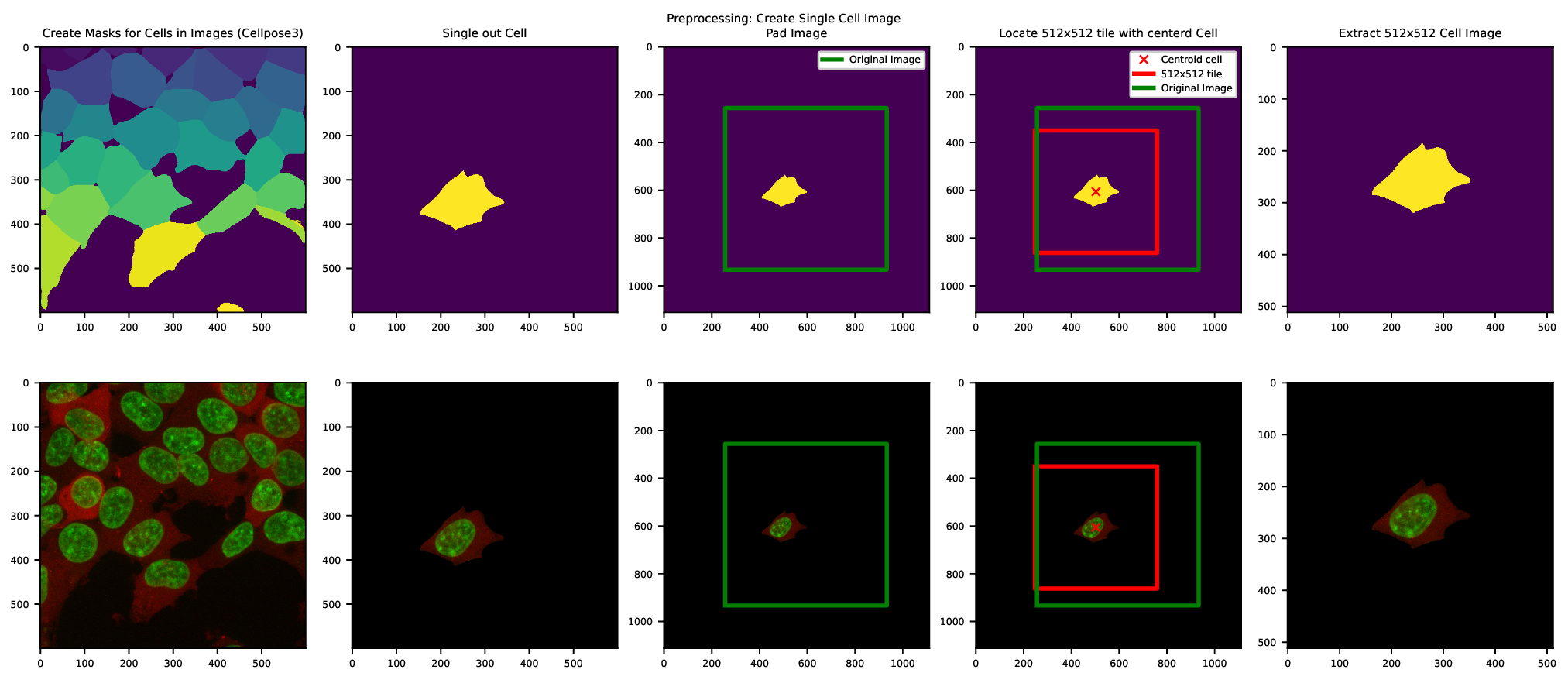}
    \caption{Extraction single-cell masks.}
    \label{fig:single_cell_crop}
\end{figure*}

\begin{figure*}[h]
    \centering
    \includegraphics[width=0.8\linewidth]{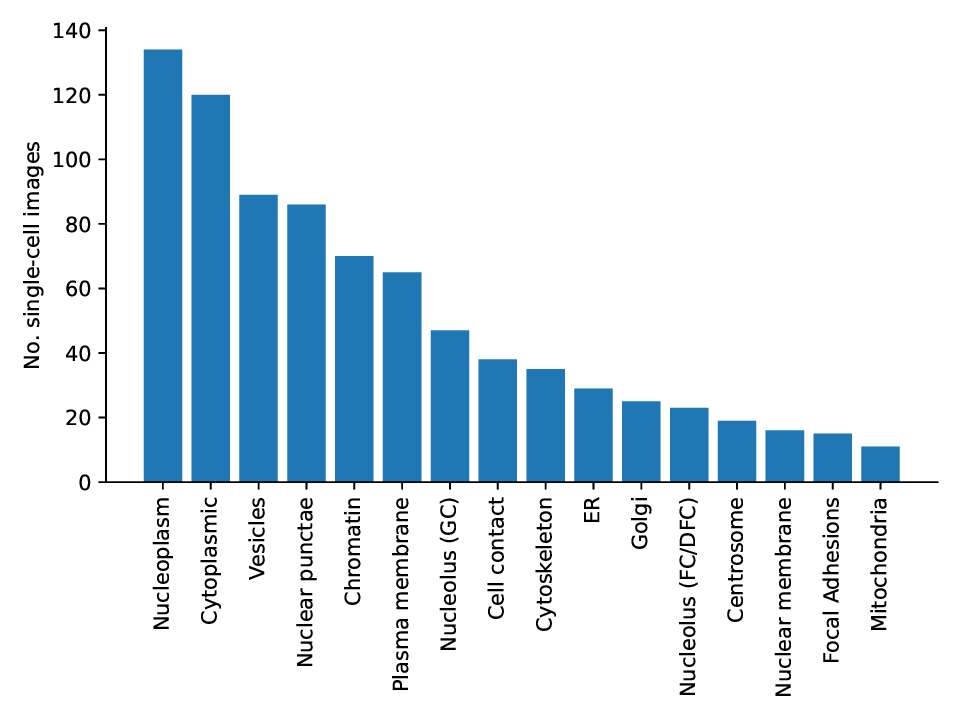}
    \caption{Number manually extracted single-cell images from the OpenCell dataset.}
    \label{fig:man_opencell_sc_images}
\end{figure*}

We applied the pretrained Cellpose v3 (cyto3) \cite{stringer_cellpose_2021,stringer_cellpose3_2025} model to each FOV and saved the instance segmentation as labeled masks. We ran Cellpose v3 with the POI as the segmentation channel and the nucleus as the additional channel. We used the default Cellpose v3 settings with automatic size estimation. Afterwards, single-cell images were obtained using the following steps (cf.~\cite{yuan_deep_2025}): Single-cell masks were detected and used to generate per-cell images. Each cell was centered using the mask centroid, and a 512$\times$512 pixel cutout was extracted around this location (Figure~\ref{fig:single_cell_crop}). 
We zoomed into the image by a factor of 2 around the center so the content appears twice as large, while keeping the image size at 512$\times$512 pixels. Cellpose v3 extracted a total of 158,762 single-cell masks from 6120 available OpenCell images. Still, the segmentation quality depended strongly on the subcellular localization of the stained protein. We observed that over-segmentation with Cellpose was common when the POI occupied only a small fraction of the cell, e.g., in images labeled "big\_aggregates", where many single-cell masks were split into multiple fragments. Since over-segmented cells are poor inputs, we applied a post-processing step to filter them out. We chose the 99.7 percent cutoff as a practical proxy for the 3$\sigma$ rule to filter images with unusually high black pixel fractions. Accordingly, we extracted 157,504 single-cell crops from the OpenCell FOV images.
We do not have the correct labels available for every single-cell image, because of the weakly-supervised nature of the OpenCell dataset. Assigning each label for every single-cell image was not feasible, therefore we generated a manually-assigned subset. We selected 61 FOV images of the OpenCell dataset and extracted single-cell masks with the Cellpose v3 model. Then, an expert reviewed the subcellular compartment labels for each single-cell mask using two criteria: (a) Does the mask capture a complete single-cell mask? (b) Does the image-level label match the single-cell label for that cell? We restricted the selection to images with Grade 3 annotations only, focusing on samples with a single label and on those representing the most common combinations of subcellular compartments in the dataset. The label "big\_aggregates" and "focal\_adhesions" were excluded because they appear in only a very small number of genes in the OpenCell dataset and are associated with unreliable segmentation. In practice, these images either suffered from over-segmentation or were dominated by more common labels, making them unsuitable for a focused and consistent analysis. From the selected 61 FOV images, we extracted 992 single-cell images and included 15 of the 17 subcellular compartments defined in OpenCell, as shown in Figure \ref{fig:man_opencell_sc_images}. 

\section{Execution Environment} \label{execution_environment}
All experiments were performed on GPU hardware. The execution environment is described in Table \ref{tab:hardware} in detail.

\begin{table}[h]
\caption{\label{tab:hardware}Hardware and Software specifications for the experimental setup.}
\begin{tabular}{ll}
\toprule 
Component & Specification \\
\midrule 
CPU & Intel Xeon Gold 6526Y, 32 CPUs (single-core sockets) @ 2.80 GHz \\
GPU & 2 $\times$ NVIDIA H100 NVL, 2 $\times$ NVIDIA L40S \\
OS & Ubuntu 22.04.5 LTS (64-bit, kernel 5.15.0-141-generic) \\
Python version & 3.10 \\
CUDA version &  11.5 \\
PyTorch version & 2.2.2 \\
\botrule 
\end{tabular}
\end{table}

\section{Impact of channel-wise embedding choices on generalization performance and finetuning}

\label{channel-embedding-appendix}

We analyze the differences and limitations of natural versus random channel-wise embeddings and assess how each strategy affects the zero-shot generalizability of pretrained backbones and the fine-tuning to OpenCell (Section~\ref{res:generalizability}). In particular, we focus on how these choices affect performance when pretrained anf fine-tuned weights are used to predict protein localization labels on OpenCell FOV images. Firstly, we examine the classification results produced by the model pretrained on ImageNet-1k when varying the embeddings, as shown in Table \ref{tab:vit-results-random-no-additional-epochs}. Here, the letters R (red), G (green), and B (blue) represent the natural pretraining channels of the ImageNet-1k backbone, while the positional list $[1, 2]$ denotes the default downstream assignment: protein at channel index 1 and nucleus at channel index 2. Alterations to this default mapping are expressed by permuting the indices; for example, $[1, 2] \rightarrow [\text{G}, \text{B}]$ represents routing the protein into the green channel and the nucleus into the blue channel.

For the OpenCell dataset, when starting from ImageNet-1k, features can only be constructed using channel-wise embedding, without natural relations between channels. This raises the question of whether performance in this arbitrarily chosen setup still depends on the specific channel distribution. We found that the choice of embedding does play a role in random channel-wise embedding. Specifically, the channel used to embed the protein channel had the strongest impact on performance. Embedding the protein channel to the green (G) channel achieved the highest results, reflected in macro $F_1$ scores of 0.697 $\pm$ 0.025 and 0.694 $\pm$ 0.029, whereas embedding the protein to the blue (B) channel resulted in the lowest performance, with macro $F_1$ scores of 0.682 $\pm$ 0.024 and 0.681 $\pm$ 0.021.
\begin{table}[h]
\caption{\label{tab:vit-results-random-no-additional-epochs}Experimental results for a ViT pretrained on ImageNet-1k evaluated on OpenCell. Values are reported as mean $\pm$ one standard deviation.}

\begin{tabular}{l l c c}
\toprule
Pretrained Dataset & Model & Channel-wise embedding & Mean Macro $F1$ \\
\midrule
ImageNet-1k & ViT-base/8 & [1, 2] $\rightarrow$ [G, B] & \textbf{0.697 $\pm$ 0.025} \\
ImageNet-1k & ViT-base/8 & [1, 2] $\rightarrow$ [G, R] & 0.694 $\pm$ 0.029 \\
ImageNet-1k & ViT-base/8 & [1, 2] $\rightarrow$ [R, G] & 0.687 $\pm$ 0.023\\
ImageNet-1k & ViT-base/8 & [1, 2] $\rightarrow$ [R, B] & 0.683 $\pm$ 0.034 \\
ImageNet-1k & ViT-base/8 & [1, 2] $\rightarrow$ [B, G] & 0.682 $\pm$ 0.024 \\
ImageNet-1k & ViT-base/8 & [1, 2] $\rightarrow$ [B, R] & 0.681 $\pm$ 0.021\\

\botrule
\end{tabular}
\end{table}
 
We further investigated whether the influence of natural channel-wise embedding persists when fine-tuning the domain-related HPA FOV backbone. As shown in Table~\ref{tab:mismatch}, random embeddings lead to a small but consistent drop ($\Delta F_1 \approx 0.01$), suggesting a minor benefit of channel-specific alignment even for domain-related backbones. For this ablation, models were trained for 60 epochs (versus 200 for the best-performing setup); the values are therefore intended for relative comparison within this ablation rather than direct comparison with the main results.

	\begin{table}[h]
	\caption{Experimental results for a ViT pretrained on the HPA FOV dataset and fine-tuned on the OpenCell dataset are reported using identical configurations across all runs. Values are reported as mean ± one standard deviation.}
	\label{tab:mismatch}
	\begin{tabular}{l l l c c}
		\toprule
		Pretrained Dataset & Model & Natural & Channel-wise embedding &  Mean Macro F1 \\
		\midrule
        HPA FOV & ViT-base/8 & Yes  & [1,2] $\rightarrow$ [1,2] & \textbf{0.702 ($\pm$ 0.022)} \\  
        HPA FOV & ViT-base/8 & No & [1,2] $\rightarrow$ [2,1] & 0.693 ($\pm$ 0.019) \\  
        HPA FOV & ViT-base/8 & No  & [1,2] $\rightarrow$ [3,0] & 0.695 ($\pm$ 0.021) \\  

	\bottomrule
\end{tabular}
\end{table}	

\section{Confusion Matrix and per-class $F_1$ score for $k$-nearest neighbor evaluation}
\label{knn-appendix}

\begin{figure}[ht]
\centering
\includegraphics[width=\textwidth]{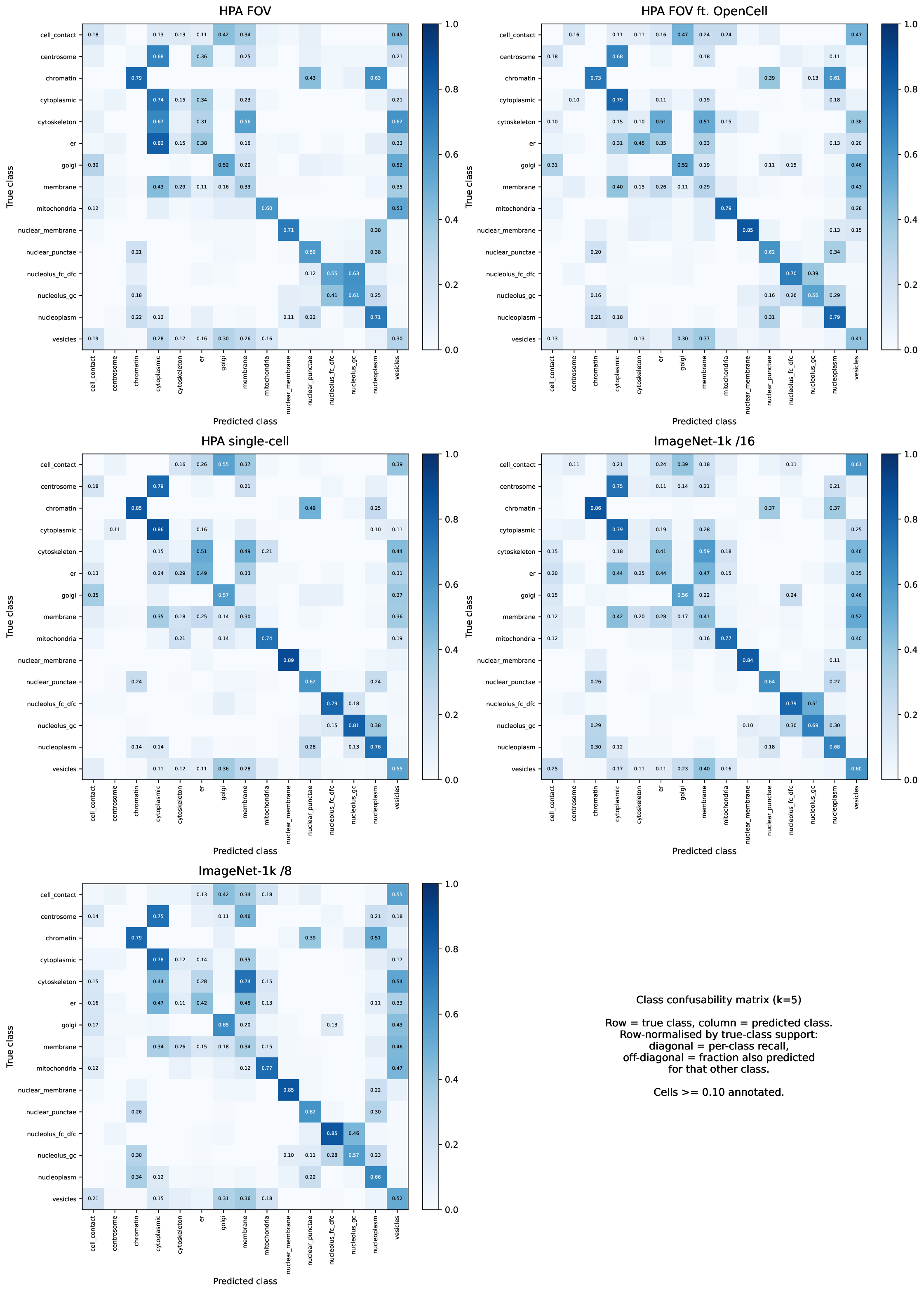}
\caption{Multi-label class confusability matrices under leave-one-out kNN
($k{=}5$) for the five zero-shot backbones. Each row is normalised by true-class
support, so the diagonal is per-class recall and off-diagonal entries give the fraction
of a true class's cells also predicted for another class.}
\label{fig:confusion-matrices}
\end{figure}

Since the mean macro $F_1$ score has limited expressiveness, on what drives the aggregate performance for the  $k$-nearest neighbor experiment, we further examine the per-class $F_1$ score and a confusion matrix to assess the influence between morphological related classes given a certain backbone. In Table \ref{tab:per-class-f1}, the aggregate advantage of the HPA single-cell backbone is not uniform across classes but is concentrated in nuclear and nucleolar localizations (nucleolus GC/FC-DFC, chromatin, nuclear membrane, nucleoplasm), where it attains the largest per-class F1 gains over the other backbones. As macro-F1 weights all classes equally, this indicates the improvement stems from sharper discrimination of these specific patterns rather than from the class-frequency distribution. The confusion matrices in Figure~\ref{fig:confusion-matrices} are largely sharing the same confusability structure across backbones — a common membrane/vesicle/Golgi cluster and mutual confusion between the two nucleolar subclasses. The HPA single-cell backbone principally sharpens the nuclear compartment: it shows the highest recall for nuclear membrane and nucleolus GC and roughly halves the cross-confusion between the two nucleolar subclasses relative to the ImageNet-pretrained backbones.

\begin{table}[ht]
\caption{Per-class F1 under leave-one-out kNN ($k{=}5$) for the five zero-shot
backbones, over all 992 annotated cells. \emph{Support} is the number of cells positive
for each class (multi-label, so columns sum to more than 992). The bottom row is the
macro-average over the 15 present classes, matching Table~\ref{tab:knn-results}. Best per
row in \textbf{bold}.}
\label{tab:per-class-f1}
\centering
\setlength{\tabcolsep}{4pt}
\begin{tabular}{l r ccccc}
\toprule
Class & Sup. & HPA FOV & FOV ft.\,OC & HPA SC & ImgNet/16 & ImgNet/8 \\
\midrule
nucleoplasm       & 184 & 0.627 & 0.702 & \textbf{0.761} & 0.668 & 0.618 \\
cytoplasmic       & 170 & 0.613 & 0.728 & \textbf{0.810} & 0.740 & 0.717 \\
nuclear\_punctae  & 120 & 0.532 & 0.509 & 0.519 & \textbf{0.611} & 0.563 \\
vesicles          & 116 & 0.227 & 0.337 & \textbf{0.462} & 0.426 & 0.369 \\
nucleolus\_gc     &  96 & 0.557 & 0.549 & \textbf{0.812} & 0.641 & 0.570 \\
membrane          &  92 & 0.271 & 0.240 & 0.271 & \textbf{0.297} & 0.230 \\
chromatin         &  84 & 0.733 & 0.709 & \textbf{0.877} & 0.713 & 0.670 \\
nucleolus\_fc\_dfc&  67 & 0.474 & 0.606 & \textbf{0.736} & 0.616 & 0.699 \\
er                &  55 & 0.249 & 0.277 & \textbf{0.360} & 0.306 & 0.348 \\
nuclear\_membrane &  55 & 0.661 & 0.770 & \textbf{0.907} & 0.786 & 0.783 \\
golgi             &  54 & 0.392 & 0.397 & 0.383 & 0.420 & \textbf{0.424} \\
mitochondria      &  43 & 0.565 & \textbf{0.654} & 0.646 & 0.559 & 0.545 \\
cytoskeleton      &  39 & 0.064 & \textbf{0.083} & 0.044 & 0.068 & 0.068 \\
cell\_contact     &  38 & \textbf{0.143} & 0.000 & 0.000 & 0.038 & 0.021 \\
centrosome        &  28 & 0.000 & 0.000 & 0.000 & 0.000 & 0.000 \\
\midrule
\textbf{Macro (present)} & & 0.407 & 0.437 & \textbf{0.506} & 0.459 & 0.442 \\
\botrule
\end{tabular}
\end{table}

\section{Extended metrics for Table 1 and 2}
\label{appendix:table1_2}
To complement the main-text results, which report mean macro $F_1$ score only, we provide
the full set of evaluation metrics for the FOV-level experiments.
Table~\ref{tab:vit-results-extended} reports macro-averaged precision ($P$), recall
($R$) and $F_1$ score, together with the micro-averaged $F_1$ score , for the zero-shot and OpenCell fine-tuning in Section~\ref{subsec:res_FOV}, computed over the same 5-fold
cross-validation pipeline described in Section~\ref{subsec:eval}. All values are
given as mean~$\pm$~standard deviation across the five gene-grouped folds.

Because the localization task is multi-label (a protein may be assigned to several
subcellular compartments simultaneously), we distinguish macro- and micro-averaging.
Macro-averaging weights all 17 classes equally regardless of their support and is
therefore the more conservative, class-imbalance-robust metric; we retain it as the
primary metric in the main text for this reason. Micro-averaging instead pools
predictions across all classes and is consequently dominated by the few
high-frequency compartments (e.g. cytoplasm, nucleoplasm), which is reflected in the
systematically higher micro-$F_1$ score  ($\approx 0.77$--$0.81$) relative to macro-$F_1$ score
($\approx 0.61$--$0.71$).

The extended metrics are consistent with the ranking reported in the main text.
Under OpenCell fine-tuning, the HPA~FOV backbone with natural channel-wise embedding
attains the highest macro $F_1$ score ($0.705 \pm 0.029$). In the zero-shot setting the
ImageNet-1k ViT-B/8 backbone with channel-wise embedding is marginally highest
($0.688 \pm 0.022$), with the HPA~FOV natural channel-wise backbone essentially on par
($0.682 \pm 0.034$); the two lie well within one standard deviation of each other. The benefit of
channel-wise embedding over channel replication is preserved in macro $F_1$ score and macro
recall across every backbone and setting; channel replication occasionally yields
marginally higher macro precision, but at a substantial cost in recall, and therefore
consistently lower $F_1$ score.

\begin{sidewaystable}[p]
\caption{\label{tab:vit-results-extended}Extended metrics on OpenCell for the
zero-shot and the OpenCell fine-tuning setting in Section \ref{subsec:res_FOV}. Macro-averaged
precision ($P$), recall ($R$) and $F1$, and micro-averaged $F1$. Values are mean
$\pm$ one standard deviation over five gene-grouped cross-validation folds; best
macro-$F1$ per setting in bold. Abbrev.: scr.\ = trained from scratch;
CW = channel-wise embedding; Nat.\ CW = natural channel-wise embedding;
Repl.\ = channel replication.}
\small
\setlength{\tabcolsep}{6pt}
\renewcommand{\arraystretch}{1.05}
\begin{tabular}{l l l c c c c}
\toprule
Pretraining & Model & Channels & Macro $P$ & Macro $R$ & Macro $F1$ & Micro $F1$ \\
\midrule
\multicolumn{7}{l}{\textit{Zero-shot (no additional fine-tuning)}} \\
\midrule
OpenCell (scr.) & ViT-S/8 & --       & 0.710 ($\pm$ 0.010) & 0.685 ($\pm$ 0.045) & 0.686 ($\pm$ 0.026) & 0.800 ($\pm$ 0.011) \\
OpenCell (scr.) & ViT-B/8 & --       & 0.654 ($\pm$ 0.020) & 0.632 ($\pm$ 0.029) & 0.631 ($\pm$ 0.021) & 0.780 ($\pm$ 0.011) \\
\cmidrule(l){1-7}
ImageNet-1k & ViT-S/8 & CW           & 0.703 ($\pm$ 0.016) & 0.659 ($\pm$ 0.025) & 0.665 ($\pm$ 0.018) & 0.786 ($\pm$ 0.009) \\
ImageNet-1k & ViT-B/8 & CW           & 0.718 ($\pm$ 0.021) & 0.679 ($\pm$ 0.032) & \textbf{0.688 ($\pm$ 0.022)} & 0.800 ($\pm$ 0.007) \\
HPA FOV     & ViT-B/8 & Nat.\ CW     & 0.708 ($\pm$ 0.022) & 0.677 ($\pm$ 0.048) & 0.682 ($\pm$ 0.034) & 0.803 ($\pm$ 0.009) \\
\cmidrule(l){1-7}
ImageNet-1k & ViT-S/8 & Repl.        & 0.707 ($\pm$ 0.023) & 0.628 ($\pm$ 0.022) & 0.653 ($\pm$ 0.018) & 0.785 ($\pm$ 0.006) \\
ImageNet-1k & ViT-B/8 & Repl.        & 0.735 ($\pm$ 0.023) & 0.645 ($\pm$ 0.028) & 0.674 ($\pm$ 0.023) & 0.803 ($\pm$ 0.005) \\
HPA FOV     & ViT-B/8 & Repl.        & 0.644 ($\pm$ 0.013) & 0.607 ($\pm$ 0.045) & 0.611 ($\pm$ 0.031) & 0.766 ($\pm$ 0.012) \\
\midrule
\multicolumn{7}{l}{\textit{OpenCell fine-tuning}} \\
\midrule
ImageNet-1k & ViT-S/8 & CW           & 0.733 ($\pm$ 0.018) & 0.696 ($\pm$ 0.046) & 0.701 ($\pm$ 0.023) & 0.809 ($\pm$ 0.006) \\
ImageNet-1k & ViT-B/8 & CW           & 0.722 ($\pm$ 0.015) & 0.691 ($\pm$ 0.042) & 0.695 ($\pm$ 0.020) & 0.807 ($\pm$ 0.011) \\
HPA FOV     & ViT-B/8 & Nat.\ CW     & 0.730 ($\pm$ 0.026) & 0.703 ($\pm$ 0.044) & \textbf{0.705 ($\pm$ 0.029)} & 0.811 ($\pm$ 0.011) \\
\cmidrule(l){1-7}
ImageNet-1k & ViT-S/8 & Repl.        & 0.692 ($\pm$ 0.018) & 0.609 ($\pm$ 0.028) & 0.633 ($\pm$ 0.020) & 0.772 ($\pm$ 0.007) \\
ImageNet-1k & ViT-B/8 & Repl.        & 0.699 ($\pm$ 0.021) & 0.612 ($\pm$ 0.037) & 0.639 ($\pm$ 0.030) & 0.766 ($\pm$ 0.013) \\
HPA FOV     & ViT-B/8 & Repl.        & 0.687 ($\pm$ 0.027) & 0.627 ($\pm$ 0.048) & 0.640 ($\pm$ 0.037) & 0.774 ($\pm$ 0.016) \\
\botrule
\end{tabular}
\end{sidewaystable}




\end{appendices}


\bibliography{sn-bibliography}

\end{document}